\documentclass[10pt,twocolumn,letterpaper]{article}

\usepackage[pagenumbers]{cvpr} 

\usepackage{graphicx}
\usepackage{amsmath}
\usepackage{amssymb}
\usepackage{booktabs}

\usepackage{times}
\usepackage{epsfig}
\usepackage{graphicx}
\usepackage{amsmath}
\usepackage{amssymb}
\usepackage{multirow}
\usepackage{algorithm2e}
\usepackage{url}
\usepackage[accsupp]{axessibility}  
\usepackage[hang,flushmargin]{footmisc}

\mathchardef\mhyphen="2D

\usepackage[pagebackref,breaklinks,colorlinks]{hyperref}

\usepackage[capitalize]{cleveref}
\crefname{section}{Sec.}{Secs.}
\Crefname{section}{Section}{Sections}
\Crefname{table}{Table}{Tables}
\crefname{table}{Tab.}{Tabs.}

\def\cvprPaperID{25} 
\def\confName{CVPR}
\def\confYear{2022}

\begin{document}


\title{Dual-Domain Image Synthesis using Segmentation-Guided GAN}

\author{Dena Bazazian $\quad\quad$ Andrew Calway $\quad\quad$ Dima Damen\\[0.5ex] Department of Computer Science, University of Bristol\\[0.5ex]
{\tt\small \{dena.bazazian, andrew.calway, dima.damen\}@bristol.ac.uk}
}
\maketitle

\begin{abstract}
We introduce a segmentation-guided approach to synthesise images that integrate features from two distinct domains. 
Images synthesised by our dual-domain model belong to one domain within the semantic-mask, and to another in the rest of the image - smoothly integrated.
We build on the successes of few-shot StyleGAN and single-shot semantic segmentation to minimise the amount of training required in utilising two domains.

The method combines few-shot cross-domain StyleGAN with a latent optimiser to achieve images containing features of two distinct domains.
We use a segmentation-guided perceptual loss, which compares both pixel-level and activations between domain-specific and dual-domain synthetic images.
Results demonstrate qualitatively and quantitatively that our model is capable of synthesising dual-domain images on a variety of objects~(faces, horses, cats, cars), domains~(natural, caricature, sketches) and part-based masks~(eyes, nose, mouth, hair, car bonnet). The code is publicly available\footnote{\scriptsize{
 \url{https://github.com/denabazazian/Dual-Domain-Synthesis}}}.
\end{abstract}


\begin{figure}
\centering
\begin{center}
\setlength{\tabcolsep}{0.001pt}
\begin{tabular}{ c c c}
\includegraphics[width=0.15\textwidth]{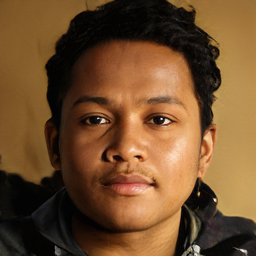}
\makebox[0pt][r]{ \raisebox{0em}{\includegraphics[width=0.030\textwidth]{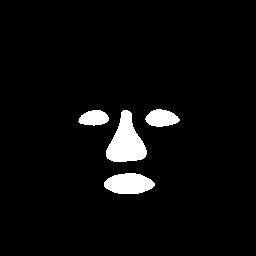} }} &
\includegraphics[width=0.15\textwidth]{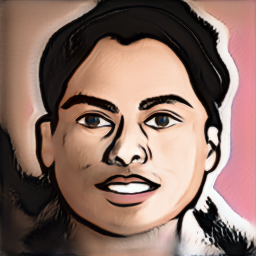}
\makebox[0pt][r]{ \raisebox{0em}{\includegraphics[width=0.030\textwidth]{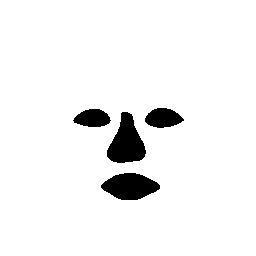} }} &
\includegraphics[width=0.15\textwidth]{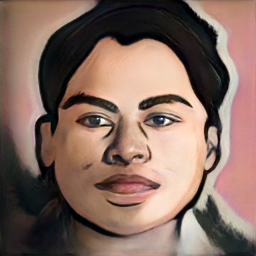}
\\
\includegraphics[width=0.15\textwidth]{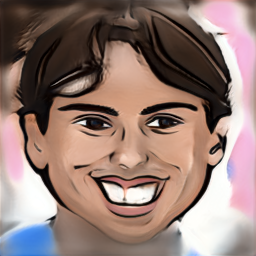}
\makebox[0pt][r]{ \raisebox{0em}{\includegraphics[width=0.030\textwidth]{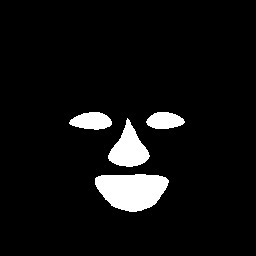} }}&
\includegraphics[width=0.15\textwidth]{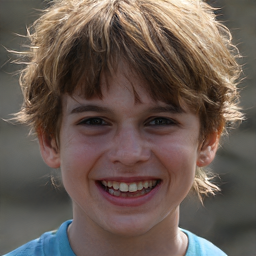}
\makebox[0pt][r]{ \raisebox{0em}{\includegraphics[width=0.030\textwidth]{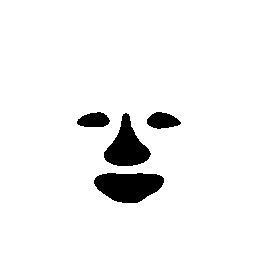} }}&
\includegraphics[width=0.15\textwidth]{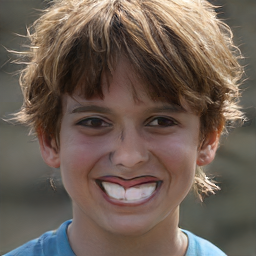}
\\
 \includegraphics[width=0.15\textwidth]{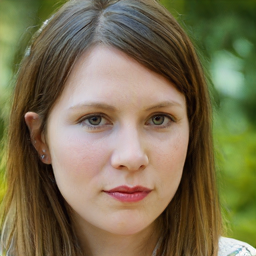}
 \makebox[0pt][r]{ \raisebox{0.0em}{\includegraphics[width=0.030\textwidth]{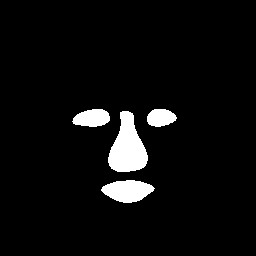} }} &

 \includegraphics[width=0.15\textwidth]{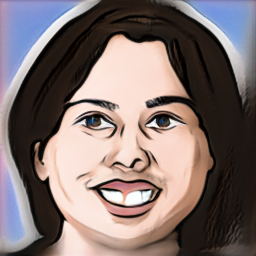}
 \makebox[0pt][r]{ \raisebox{0.0em}{\includegraphics[width=0.030\textwidth]{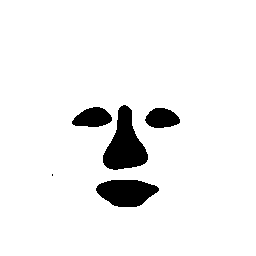} }} &
\includegraphics[width=0.15\textwidth]{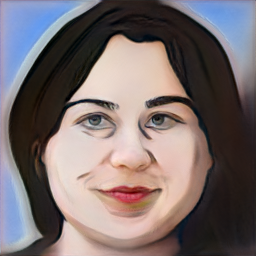}
\end{tabular}
\end{center}
\vspace*{-21pt}
\caption{Dual-Domain Synthesis combines segmentation-guided features from the source domain (left), and the target domain (centre) - corresponding segmentation masks are incorporated.
Synthesised images (right) -- top\&bottom: caricature face with natural eyes/nose/mouth, middle: natural face with caricature features.
The first two rows are from paired examples, i.e. the same latent code, while the third row shows an unpaired example.}
\label{fig:introduction}
\end{figure}

\section{Introduction}

Automated image manipulation, particularly portrait editing, is of significant interest to researchers and practitioners alike with applications in animation, gaming and social media entertainment.  
Advances in Generative Adversarial Networks (GANs)~\cite{GAN} have led to tremendous progress and innovation, enabling the synthesis of realistic images~\cite{stylegan1,stylegan2}, and leading to techniques for image-to-image translation~\cite{image_translation} and domain translation~\cite{fewShotGAN}, able to synthesise a corresponding image of one domain using a reference image from another, transferring the style.

In this paper, we focus on segmentation-guided dual-domain feature integration -- combining segmentation parts from one domain with an image from another in a way that is visually convincing. For example, replacing the eyes and mouth of a natural face image with those from a caricature, so that the result is a dual-domain image. Examples are shown in Fig.~\ref{fig:introduction}. 
Note the distinction from previous mask-guided transfer methods which combine styles and/or features from the same domain~\cite{SEAN2020,SPADE,portrait_maskguided}.

We utilise the few-shotGAN~\cite{fewShotGAN} technique to generate related images of two domains from a common latent space.
To create the segmentation masks efficiently, we employ the one-shot segmentation approach of RepurposingGAN~\cite{repurposingGAN}. 
We leverage these two StyleGAN-based approaches~\cite{fewShotGAN,repurposingGAN} to obtain input images and masks, and propose a segmentation-guided weighted combination with perceptual and pixel-wise losses to synthesise the dual-domain image.

To the best of our knowledge, this is the first attempt to synthesise images based on the features from two domains. 
The contributions can be summarised as follows. First, we introduce Dual-Domain Image Synthesis, a technique for generating images by integrating the features of two domains, guided by a semantic segmentation model.  
Second, we demonstrate that the approach works for images of faces, horses, cats and cars, synthesising dual-domain images from natural, caricature and sketch images. We test a variety of part-based segmentations for guidance, such as hair in faces, eyes in cats, heads of horses and car bonnets.



\section{Related work}
\noindent \textbf{Generative Models:}
Deep generative models have demonstrated robust results in generating realistic images for various tasks such as image segmentation~\cite{repurposingGAN,cycleGAN_segmentation,text_segmentation,forground_segmentation}, image inpainting~\cite{PDGAN}, inverse graphics~\cite{InverseGraphics2020}, conditional image synthesising~\cite{SEAN2020,SPADE} and domain translations~\cite{fewShotGAN}.
State-of-the-art GANs, such as StyleGAN~\cite{stylegan1} and StyleGAN2~\cite{stylegan2}, are capable of generating incredibly realistic images at a high resolution, making these popular models in works such as~\cite{repurposingGAN,InverseGraphics2020,fewShotGAN,image2stylegan++,image2stylegan,few-shot-adaptation}. 
The manifolds of changes in spatially localised regions for the images generated by StyleGAN are explored in~\cite{maskGuided_Manifold}.  
Cross-domain correspondence techniques based on StyleGAN are proposed in~\cite{fewShotGAN,few-shot-adaptation } to generate images from two domains using a handful of examples. 
We make use of ~\cite{fewShotGAN} since it demonstrates better performance on far-domain adaptation. In this work we instead generate images which integrate the two domains into a single image.

\noindent \textbf{Latent Optimisation:}
One of the powerful aspects of StyleGAN-based methods is the ability to generate realistic images from random latent codes and hence the option of seeking optimal codes w.r.t chosen loss criteria. A number of recent papers have adopted this approach to generate images based on perceptual loss, by comparing activation layers from convolutional networks~\cite{perceptual_loss, unreasonable_perceptual_metric, image2stylegan, image2stylegan++}.
Optimising for the perceptual loss in~\cite{perceptual_loss} allows transferring the style from one image to another. Furthermore, perceptual loss is used in~\cite{image2stylegan,image2stylegan++}
to fit a projected synthetic image into the distribution of images of a particular domain.

\noindent \textbf{Image Completion:}
We here group works that attempt mask-guided image synthesis and inpainting.
Mask-guided synthesise techniques combine source and reference images based on a given mask to generate a realistic image~\cite{SEAN2020,SPADE,portrait_maskguided}, while image inpainting techniques fill an arbitrary hole of a given image to generate a natural image~\cite{PDGAN, imageInpaintingIrregular}. Part-based editing in synthetic images was proposed in~\cite{LatentStyleEdit} to transfer the appearance of a specific object part from a reference image to a target image.  K-means is applied to the hidden layer activations of the StyleGAN generator to reveal a decomposition of the generated output into semantic objects and object-parts.
However, note that all these previous part-based methods combine images from the same domain, contrasting with our approach which seeks to integrate features from distinct domains.

\noindent \textbf{Image Blending:} Image blending aims to seamlessly blend an object from a source image onto a target image with lightly mask adjustment~\cite{zhang2020deep}.
\cite{wu2019gp} combines synthesis with Gaussian-Poisson blending, and is trained using blended ground-truth. \cite{zhang2020deep} combines a Poisson gradient loss, style loss, and content loss, without the need for training data.
Similarly, image Harmonisation adjusts the illumination, colour, and texture of the foreground mask. 
All blending approaches aim to general in order to apply to any pair of images.
In contrast, our proposed approach integrates semantic parts from one domain, importantly modifying the full image to ensure it belongs to a specific target domain.

\section{Proposed Approach }
We aim to generate a visually convincing image that integrates features of two domains, guided by part-based segmentations.
Our pipeline utilises two generative models for the source and target domains along with a segmentation model for localising the corresponding mask to integrate the features of two domains. 
The masks indicate the semantic parts of the image that should be integrated from one domain into the other. 
We start by revisiting works we build on, then detail our approach. 


\subsection{Revisiting Few-shotGAN}
\label{sec:fewshot}
The work of few-shotGAN~\cite{fewShotGAN} considers a main domain, for which a large-scale dataset is readily available, and trains a StyleGAN model on this data. 
Given few-shot examples (e.g.\ ten samples) from a second domain, 
few-shotGAN~\cite{fewShotGAN} utilises the StyleGAN trained on the large  dataset for pretraining and transfers the  diversity information from this model to a second model. The training preserves the relative similarities and  differences between instances via a cross-domain distance consistency loss. 

We use~\cite{fewShotGAN} to train two generative models, one with many-shot examples and the other adapted from the first with few-shot examples. 
Importantly, the two generative models $\mathcal{G}_1$ and $\mathcal{G}_2$ share the same latent space $\mathcal{Z}$ during adaptation.
This is of particular importance to our method, as we base the ability to integrate features from both domains on sampling from this common latent space.
Given the same random vector $ z^\star \subset \mathcal{Z}$, two images are synthesised, one from each generator, and thus from two domains.
We refer to these as \emph{paired} images as they share the same latent code, albeit from two domains.

\subsection{Revisiting RepurposingGAN}
\label{sec:repurposingGAN}
We train a part-based semantic segmentation model, using the method from RepurposingGAN~\cite{repurposingGAN}.
Each synthesised image is passed through a series of spatial convolutions, and a binary pixel-level segmentation is computed from a unique generative computation which can be traced back through each convolutional layer down to the initial latent code. The strength of~\cite{repurposingGAN} is based on the capability to train this segmentation model with one-shot, i.e. by manually segmenting a single synthetic image.

We use RepurposingGAN~\cite{repurposingGAN} to train our segmentation model from a randomly synthesised image of one domain.
The same segmentation model is used for semantically segmenting synthetic images from both domains.
This is only plausible thanks to the adaptation method in Sec.~\ref{sec:fewshot}.
We demonstrate experimentally that this method is robust across domains for a variety of semantic masks.

\subsection{Dual-Domain Synthesis (DDS)}
We introduce Dual-Domain Synthesis (DDS) framework in this section. We consider as input to our network two generative models from two domains (Sec.~\ref{sec:fewshot}).
We select one of these two domains to be the `source' domain ($\mathcal{D}_s$), and the other as the `target' domain ($\mathcal{D}_t$)\footnote{Note that the source domain may or may not correspond to the few-shot domain in the adaptation, and similarly for the target domain.}, and accordingly label the two trained generative models as $\mathcal{G}_s$ and $\mathcal{G}_t$.
Given one latent vector $z^\star \in \mathcal{N}(\mathbf{0},\mathbf{1})$, we generate two synthetic images such that
\( x_s \in \mathbb{R}^{n \times n \times 3} \), where \(x_s\) is an image from the source domain ($\mathcal{D}_s$), and \( x_t \in \mathbb{R}^{n \times n \times 3} \), where \(x_t\) is an image from the target domain~($\mathcal{D}_t$). 
We pass both images to the trained semantic segmentation model (Sec.~\ref{sec:repurposingGAN}).
Corresponding spatial masks denoting the semantic part in the source and target images are defined as $y_s$ and~$y_t$.
Our goal is to synthesise one image, that is perceptually close to the \emph{target} domain, but contains features from the \emph{source} domain.
The spatial masks are used for guiding the regions that should be transferred from source to target.

\begin{figure}[t]
\begin{center}
\includegraphics[width=1\linewidth]{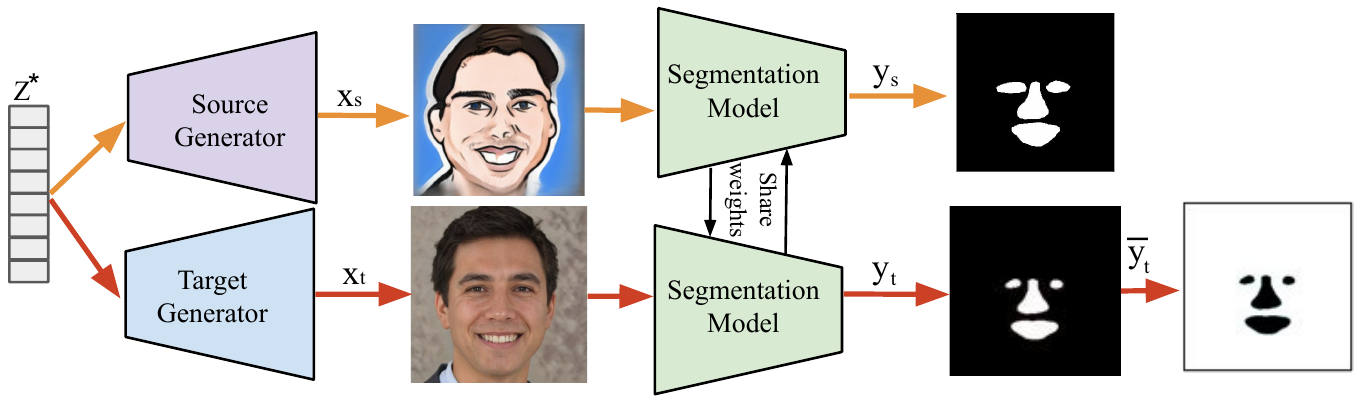}\\    \includegraphics[width=1\linewidth]{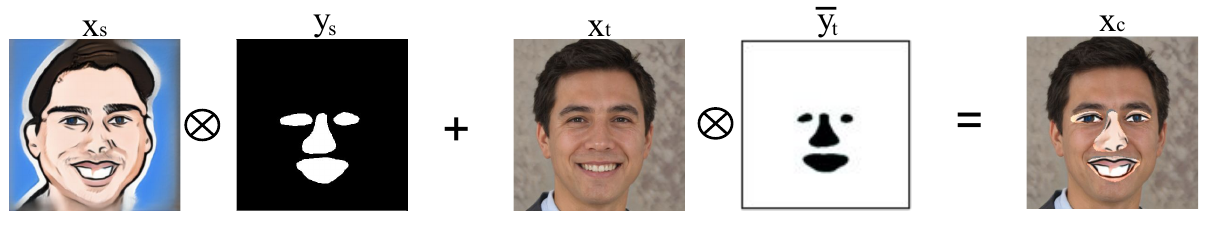}
\end{center}
\vspace*{-21pt}
   \caption{\textbf{Top: } A random latent code $z^\star$ is fed into two different generators to create two corresponding images from two different domains, source image $x_s$ and target image $x_t$. Masks of the generated images, $y_s$ and $y_t$, are then computed from a one-shot segmentation model. 
   \textbf{Bottom: }Naive crossover image~($x_c$), where $\bar{y}_t$ denotes the binary complement of $y_t$. }
\label{fig:generate_image_mask}
\end{figure}

A naive approach to achieve dual-domain synthesis is simply to use additive `cut and paste' to combine the relevant masked regions. We denote this as the crossover image~($x_c$).
The source image $x_s$ is masked by the corresponding regions in $y_s$ indicating the parts we wish to integrate from the source domain.
Additionally, we wish to remove/subtract the corresponding segmentation parts from the target image. We thus use the binary complement of the target mask as $\bar{y}_t$. 
Fig.~\ref{fig:generate_image_mask} shows examples of the generated source and target images and their corresponding masks.
The naive crossover image $x_c$ is thus computed as:   
\begin{equation}
    x_c = (x_s \otimes y_s) + (x_t \otimes \bar{y}_t),
\label{eq:naive_crossover} 
\end{equation}
where $\otimes$ is pixel-wise multiplication/masking.



Importantly, while this image contains the features from both domains, it is not a visually convincing dual-domain image. 
We thus aim to search the latent space $\mathcal{Z}$ for the the optimal latent vector $\hat{z}$ that can generate our dual-domain image such that  ${\hat{x}_t \gets \mathcal{G}_t(\hat{z})}$.
We use a gradient-based optimisation that iteratively updates a latent code ${\hat{z} \in \mathcal{N}(\mathbf{0},\mathbf{1})}$ using a loss function that is based on a combination of perceptual loss~\cite{perceptual_loss}
and pixel-level Mean Square Error (MSE). 

We initialise $\hat{z}$ randomly\footnote{Initial experimental evidence showed that starting from $z^\star$ produces inferior results with the optimisation stuck in local minima.}.
We base our perceptual loss on that proposed in \cite{image2stylegan}, which compares the activations of a pre-trained convolutional network from two images, a synthesised image and a reference image.
In DDS, we do not have a reference image in the dual-domain, and thus our perceptual loss considers both the source and target domains.

\begin{figure*}[t]
\begin{center}
\includegraphics[width=0.95\linewidth,height=9cm]{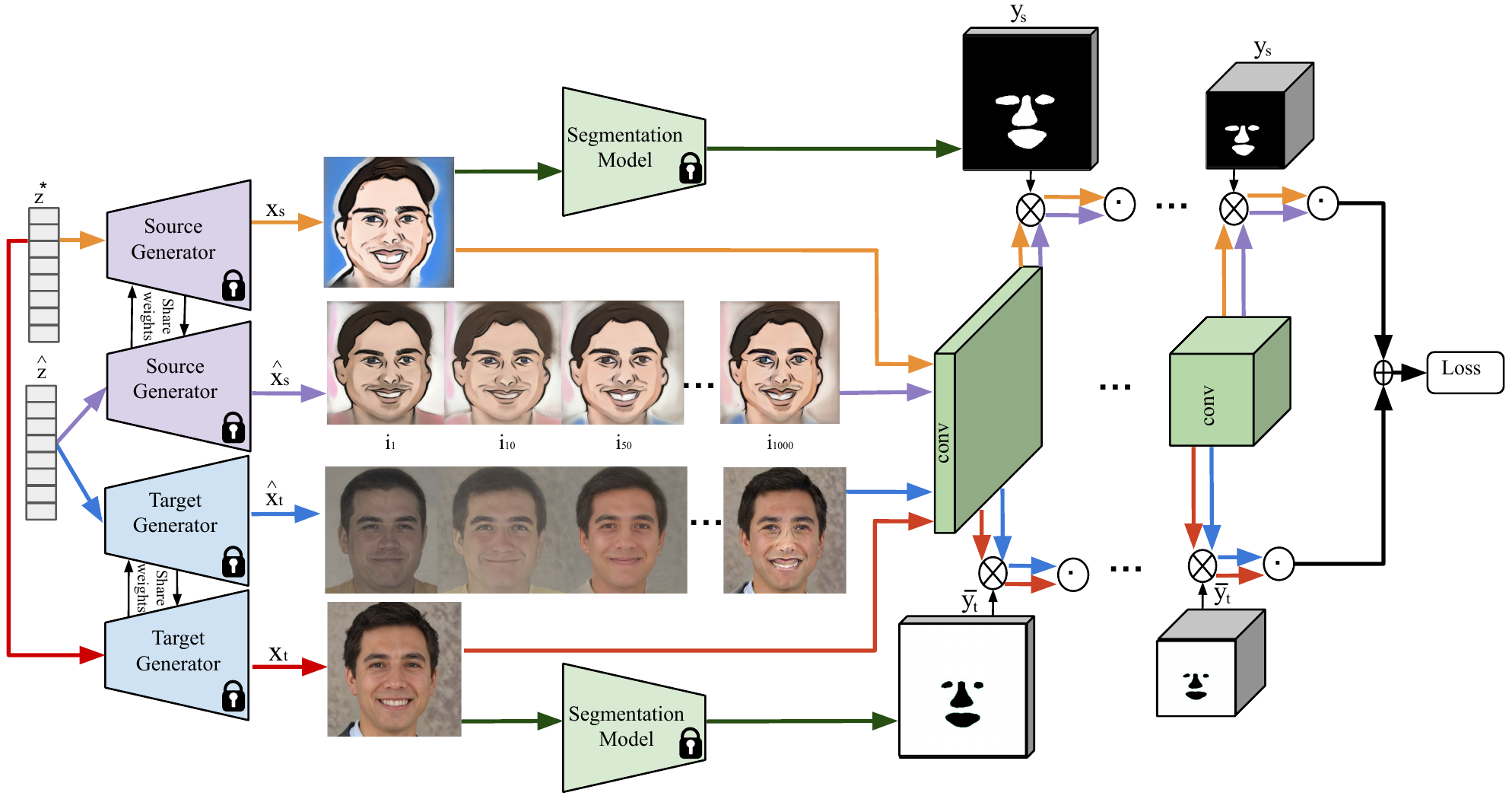}
\end{center}
\vspace*{-15pt}
\caption{\textbf{Perceptual latent loss optimisation.} $ \odot , \otimes $ and $\oplus $ indicate MSE loss, multiplication and summation, respectively. The colours of each arrow depict distinct data paths. $x_s, y_s, x_t$ and $\bar{y}_t$ match those illustrated in Fig.\ref{fig:generate_image_mask}. $z^\star$ is fixed while $\hat{z}$ is a random latent vector that is to be optimised. $\hat{x}_s$ and $\hat{x}_t$ are generated iteratively based on the optimisation of $\hat{z}$. Examples of $\hat{x}_s$ and $\hat{x}_t$ are shown for different iterations during the optimisation ($i_1, i_{10}, i_{50}, i_{1000}$).} 
   \vspace*{-12pt}
\label{fig:crossover}
\end{figure*}

We use activations from the layers $conv_{1\_1}$, $conv_{1\_2}$, $conv_{2\_2}$ and $conv_{3\_3}$ of \emph{VGG-16}~\cite{VGG} for the perceptual loss as proposed in~\cite{image2stylegan}. 
We downsample the segmentation binary masks at each layer to match the resolutions of the corresponding activation layer, and duplicate the mask across the channel dimension, to match the number of channels in the layer. 
To measure the similarity between the generated image and the expected image during optimisation, we apply a loss function that is a weighted combination of perceptual loss and the pixel-wise MSE loss:
\begin{equation}
\begin{split}
    &\mathcal{L}_s = \mathcal{L}_p(\mathcal{G}_s(\hat{z}), x_s,y_s) + \left \| \mathcal{G}_s(\hat{z}) \otimes y_s - x_s \otimes  y_s \right \|^2  \\
    &\mathcal{L}_t = \mathcal{L}_p(\mathcal{G}_t(\hat{z}), x_t,\bar{y}_t) + \left \| \mathcal{G}_t(\hat{z}) \otimes \bar{y}_t  - x_t \otimes  \bar{y}_t \right \|^2
\end{split}
\label{eq:l}
\end{equation}
where $x_s, x_t$ are source and target images, $\bar{y}_t$ is a binary complement of the target mask $y_t$. $\mathcal{G}_s(\cdot), \mathcal{G}_t(\cdot)$ are the generators for source and target domains respectively.
The perceptual loss $\mathcal{L}_p(\cdot)$ in Eq.~\ref{eq:l} is computed as :
\begin{equation}
\mathcal{L}_p (I_1, I_2,\lambda)=
\sum_{j=1}^{J}\frac{1}{N_j}\left \| \mathcal{F}_j (I_1) \times \lambda_j -  \mathcal{F}_j(I_2) \times \lambda_j  \right \|
\label{eq:perceptual_loss}
\end{equation}
where $I_1 , I_2$ are two images, $\lambda$ is the mask, $\lambda_j$ is the downsampled segmentation mask according to the size of each $conv$ layer $j$.
$\mathcal{F}_j$ is the activations of the $conv$ layers and $N_j$ is the number of scalars in the $j^{th}$ layer output, and $J$ is the number of convoultional layers.
Fig.~\ref{fig:crossover} shows an illustration of the process of the perceptual loss. 

The perceptual loss is combined with the pixel-wise MSE loss in Eq.~\ref{eq:l} for integrating the features from the two domains.
In order to increase the quality of the generated image, and ensure its segmentation-guided similarity to the source and target images, we use an additional MSE loss based on the naive cross over image as: 
\begin{equation}
\mathcal{L}_c =  \left \| s_c-x_c\right \|
\label{eq:naive_crossover_LOSS}
\end{equation}
where $x_c$ is the naive crossover image from Eq.~\ref{eq:naive_crossover}, and $s_c$ is a naive crossover synthetic image computed each iteration $i$ as:
\begin{equation}
    s_c = (\mathcal{G}_s(\hat{z}) \otimes y_s) + (\mathcal{G}_t(\hat{z}) \otimes \bar{y}_t).
\label{eq:naive_crossover_synthetic}    
\end{equation}
The overall loss is then defined as:
\begin{equation}
\mathcal{L}oss = \alpha \mathcal{L}_s + \beta \mathcal{L}_t + \gamma \mathcal{L}_c,
\label{eq:loss}
\end{equation}
Note that we only optimise $\hat{z}$ and do not update the weights of the convolutional network, generative networks or the domain-specific synthetic images $x_s, x_t$.
We then use the target domain generator to produce the dual-domain image, such as: $ \hat{x} \leftarrow \mathcal{G}_t (\hat{z})$. The complete process is summarised in Algorithm~\ref{alg:DBlender}. 

\RestyleAlgo{ruled}
\SetKwInOut{Parameter}{Parameter}
\SetKwInOut{Model}{Model}

\begin{algorithm}[t!]
\caption{Dual-Domain Synthesis algorithm} \label{alg:DBlender}

\KwIn{$x_s, x_t, y_s ,\bar{y}_t$}
\Model{$\mathcal{G}_s,\mathcal{G}_t$}
\Parameter{$\alpha, \beta, \gamma$}
\KwResult{$\hat{x}$} 
$x_c \gets x_s \otimes y_s + x_t \otimes \bar{y}_t$\;
$\hat{z} \gets \mathcal{N}(\mathbf{0}, \mathbf{1})$\; 
$\nabla \gets Optimiser$\;

\While{i $<$ maxIterations}{
$\mathcal{L}_s, \mathcal{L}_t  \gets Eq.\ref{eq:l} $\;
$\mathcal{L}_c \gets  Eq.\ref{eq:naive_crossover_LOSS}$\;
$\mathcal{L}oss \gets \alpha \mathcal{L}_s + \beta \mathcal{L}_t + \gamma \mathcal{L}_c$\; 
$ \hat{z} \gets \nabla(\hat{z}) $
}
$\hat{x} \gets \mathcal{G}_t(\hat{z})$\;
\end{algorithm}


 \begin{figure*}[t]
\centering
\begin{center}
\setlength{\tabcolsep}{0.5pt}
\begin{tabular}{ c c c c c c c c c c c | c c }
$\hat{x}_{t_{1}}$ &  $\hat{x}_{t_{5}}$ & $\hat{x}_{t_{10}}$ & $\hat{x}_{t_{15}}$ & $\hat{x}_{t_{20}}$ & $\hat{x}_{t_{25}}$ & $\hat{x}_{t_{30}}$ & $\dots$ & $\hat{x}_{t_{500}}$ & $\dots$ & $\hat{x}_{t_{1000}}$ & $x_t$ & $x_s$ 
\\
\includegraphics[width=0.083\textwidth]{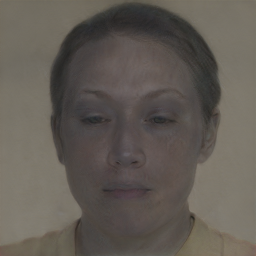} &
\includegraphics[width=0.083\textwidth]{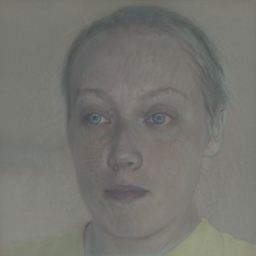} &
\includegraphics[width=0.083\textwidth]{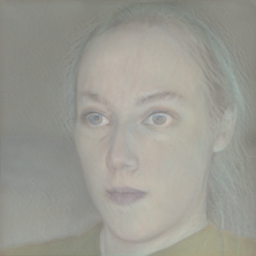} &
\includegraphics[width=0.083\textwidth]{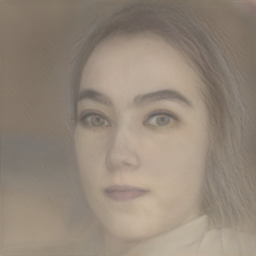} &
\includegraphics[width=0.083\textwidth]{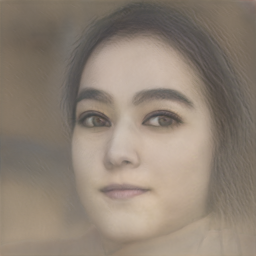} &
\includegraphics[width=0.083\textwidth]{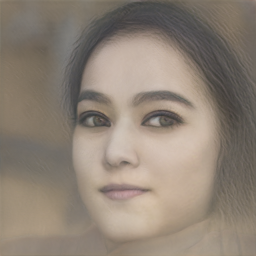} &
\includegraphics[width=0.083\textwidth]{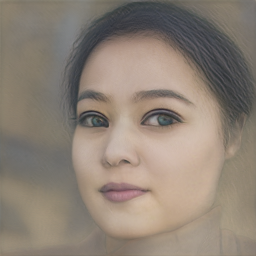} &
$\dots$ &
\includegraphics[width=0.083\textwidth]{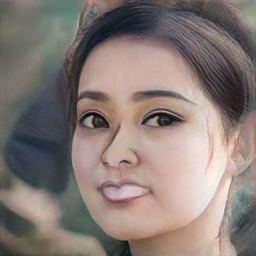} &
$\dots$ &
\includegraphics[width=0.083\textwidth]{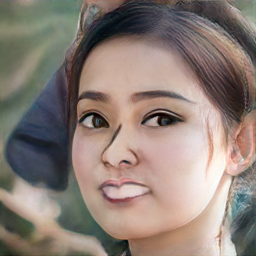} &
\includegraphics[width=0.083\textwidth]{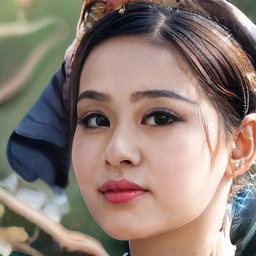} &
\includegraphics[width=0.083\textwidth]{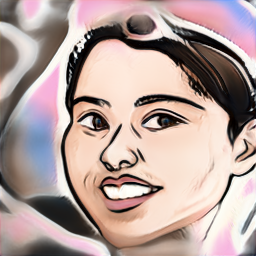} 
\\
\includegraphics[width=0.083\textwidth]{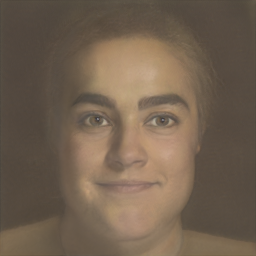} &
\includegraphics[width=0.083\textwidth]{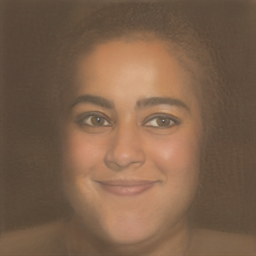} &
\includegraphics[width=0.083\textwidth]{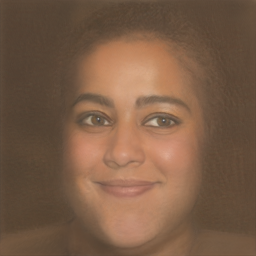} &
\includegraphics[width=0.083\textwidth]{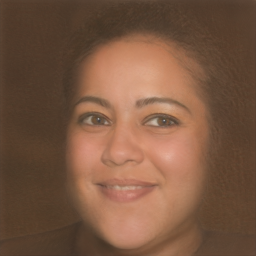} &
\includegraphics[width=0.083\textwidth]{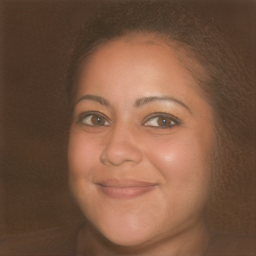} &
\includegraphics[width=0.083\textwidth]{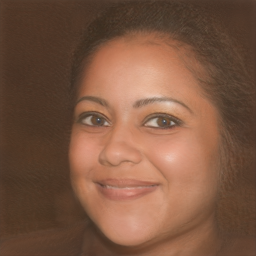} &
\includegraphics[width=0.083\textwidth]{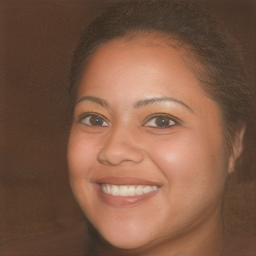} &
$\dots$ &
\includegraphics[width=0.083\textwidth]{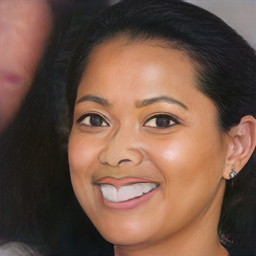} &
$\dots$ &
\includegraphics[width=0.083\textwidth]{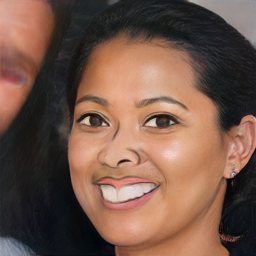} &
\includegraphics[width=0.083\textwidth]{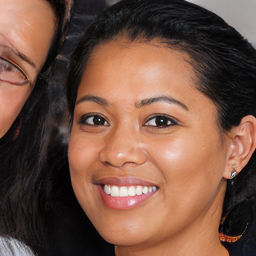} &
\includegraphics[width=0.083\textwidth]{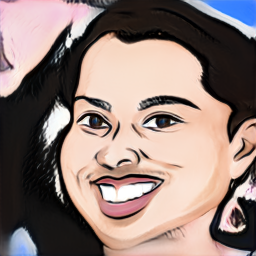} 
\\
\includegraphics[width=0.083\textwidth]{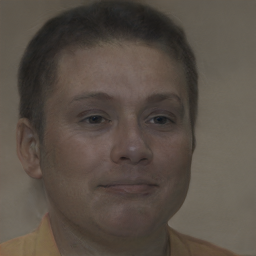} &
\includegraphics[width=0.083\textwidth]{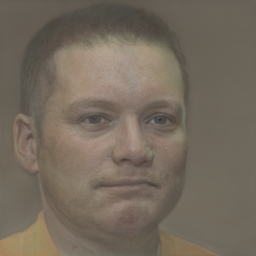} &
\includegraphics[width=0.083\textwidth]{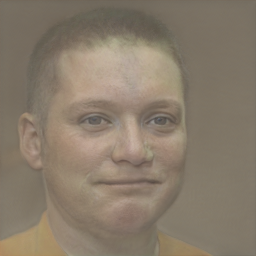} &
\includegraphics[width=0.083\textwidth]{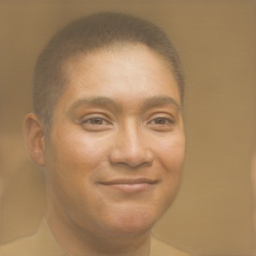} &
\includegraphics[width=0.083\textwidth]{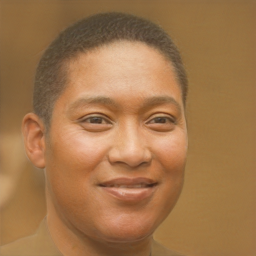} &
\includegraphics[width=0.083\textwidth]{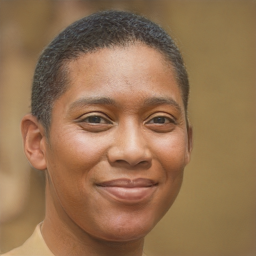} &
\includegraphics[width=0.083\textwidth]{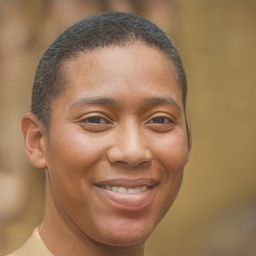} &
$\dots$ &
\includegraphics[width=0.083\textwidth]{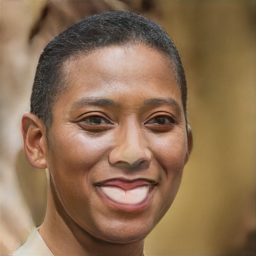} &
$\dots$ &
\includegraphics[width=0.083\textwidth]{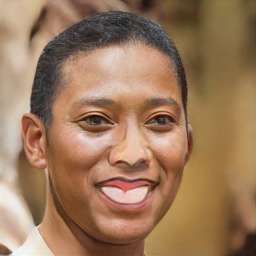} &
\includegraphics[width=0.083\textwidth]{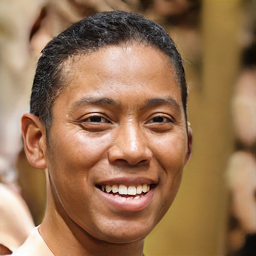} &
\includegraphics[width=0.083\textwidth]{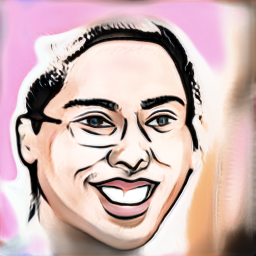} 
\\
\includegraphics[width=0.083\textwidth]{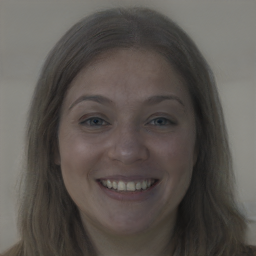} &
\includegraphics[width=0.083\textwidth]{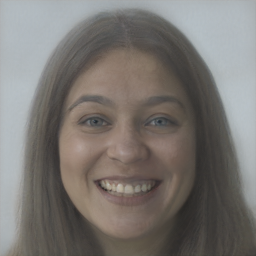} &
\includegraphics[width=0.083\textwidth]{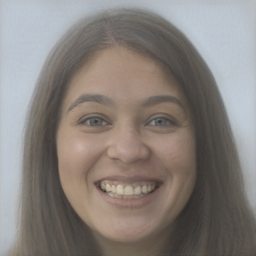} &
\includegraphics[width=0.083\textwidth]{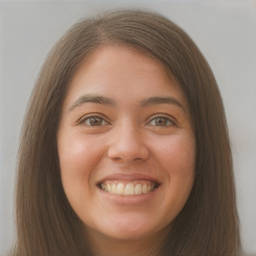} &
\includegraphics[width=0.083\textwidth]{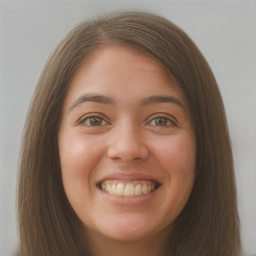} &
\includegraphics[width=0.083\textwidth]{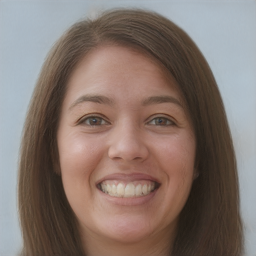} &
\includegraphics[width=0.083\textwidth]{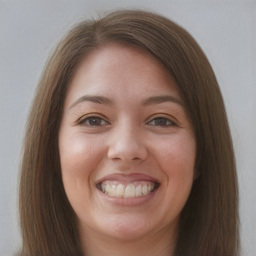} &
$\dots$ &
\includegraphics[width=0.083\textwidth]{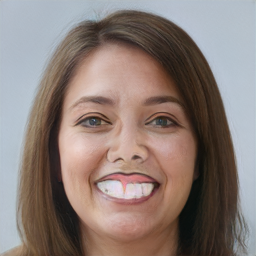} &
$\dots$ &
\includegraphics[width=0.083\textwidth]{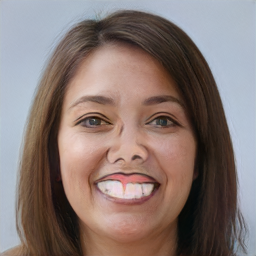} &
\includegraphics[width=0.083\textwidth]{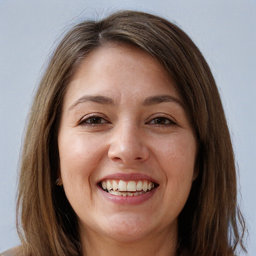} &
\includegraphics[width=0.083\textwidth]{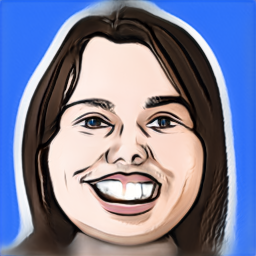}
\end{tabular}
\end{center}
\vspace*{-20pt}
\caption{Intermediate DDS results for $DDS_{ car.\rightarrow nat.}$ with a semantic mask of the eye/nose/mouth regions. Source and target domains are caricature and natural faces respectively. In the first row, note the distinct caricature eyes, nose and mouth in $i \ge 500$ iterations. } 
\label{fig:iterations}
\end{figure*}
\section{Implementation Details}


We used the StyleGAN2 and few-shotGAN implementations provided in \cite{stylegan_github} and \cite{fewshotGAN_github}, and 10 images for few-shot far-domain adaptation in each case.
Segmentation models are based on the implementation of RepurposingGAN provided in \cite{repurposeGAN_github}. 
We manually label semantic masks in each case as follows.
We first synthesise an image by feeding a random latent code to a StyleGAN model~\cite{stylegan2}.  
We then manually label the mask of the part(s) of interest, and train the segmentation model as in \cite{repurposeGAN_github}.

In all experiments, we optimise DDS for $1000$ iterations, using a learning rate of $0.01$ with the Adam optimiser~\cite{adam}.
$J$ in Eq.~\ref{eq:perceptual_loss} is set to 4, given the layers chosen from \emph{VGG-16}.
Based on empirical experiments we set the $ {\alpha , \beta , \gamma } $ in Eq.~\ref{eq:loss} as ${0.9,1,0.5}$, respectively. 
Robustness to these hyperparameters was evaluated using an ablation study in Sec.~\ref{sec:exp_face}.

\section{Experiments}
In this section, we present and discuss quantitative and qualitative experimental results of DDS on the domains of face, horse, car and cat images.


\subsection{DDS on Face Domains}  
\label{sec:exp_face}
We perform experiments on human face images based on a StyleGAN model~\cite{stylegan_github} trained on Flickr-Faces-HQ~(FFHQ)~\cite{stylegan2} to generate natural human face images. Then, we use few-shotGAN model trained on Face-Caricature~\cite{fewShotGAN} and Face-Sketch~\cite{sketch_face} images (models  from~\cite{fewshotGAN_github}). 
We experiment with the domain pairs natural~$\leftrightarrow$~caricature and natural $\leftrightarrow$ sketch and two part-based segmentation masks, one indicating the regions of eyes, nose and mouth\footnote{We combine three face parts into one mask (nose, eyes and mouth) to show more experimental results in one image.} and the other indicating the hair.

\noindent\textbf{Qualitative results:}
Fig.~\ref{fig:iterations} first shows the intermediate results of DDS on the experiments for integrating the features of caricature domain into the natural face domain based on the eye/nose/mouth segmentation model. We illustrate the evolution of $\hat{x}_{t_{i}}$ while the latent code $\hat{z}$ is being iteratively optimised to generate the dual-domain image.
We compare these to the target and source domains.
Note that the optimisation seems to first focus on the identity and pose in the target domain, with the last iterations focusing on the dual-domain image synthesis.

\begin{figure*}
\centering
\begin{center}
\setlength{\tabcolsep}{0.01pt}
\begin{tabular}{ c c | c c | c c | c c }

\includegraphics[width=0.12\textwidth]{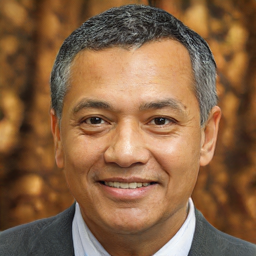} 
\makebox[0pt][r]{ \raisebox{0.0em}{\includegraphics[width=0.030\textwidth]{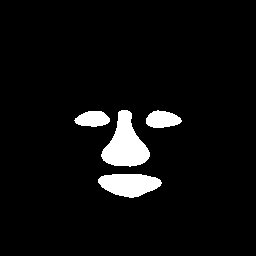} }} &
\includegraphics[width=0.12\textwidth]{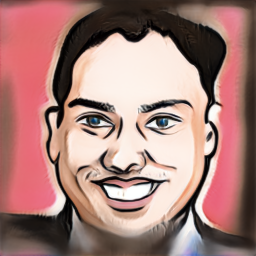} 
\makebox[0pt][r]{ \raisebox{0.0em}{\includegraphics[width=0.030\textwidth]{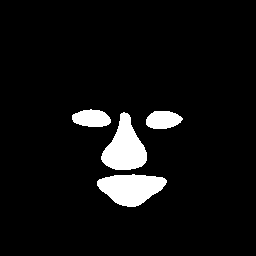} }} &
\includegraphics[width=0.12\textwidth]{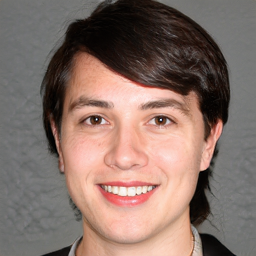} 
\makebox[0pt][r]{ \raisebox{0.0em}{\includegraphics[width=0.030\textwidth]{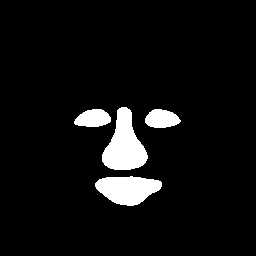} }} &
\includegraphics[width=0.12\textwidth]{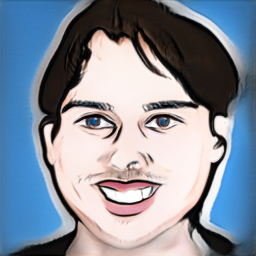} 
\makebox[0pt][r]{ \raisebox{0.0em}{\includegraphics[width=0.030\textwidth]{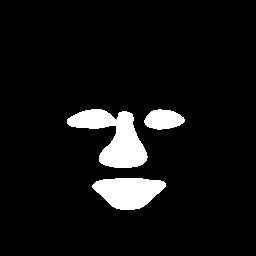} }}&
\includegraphics[width=0.12\textwidth]{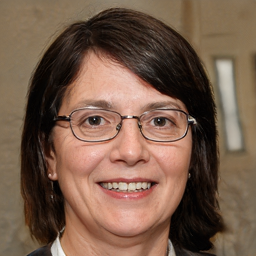} 
\makebox[0pt][r]{ \raisebox{0.0em}{\includegraphics[width=0.030\textwidth]{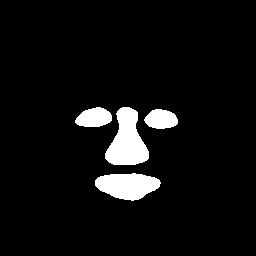} }} &
\includegraphics[width=0.12\textwidth]{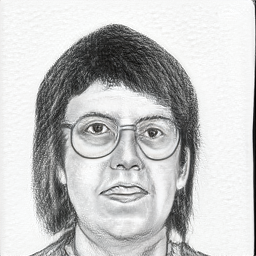} 
\makebox[0pt][r]{ \raisebox{0.0em}{\includegraphics[width=0.030\textwidth]{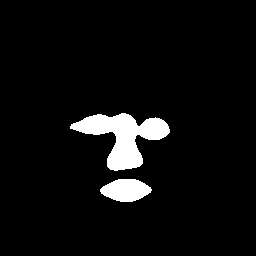} }} &
\includegraphics[width=0.12\textwidth]{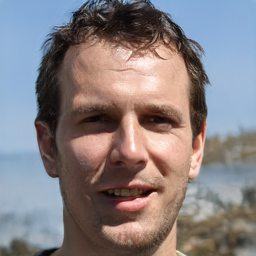} 
\makebox[0pt][r]{ \raisebox{0.0em}{\includegraphics[width=0.030\textwidth]{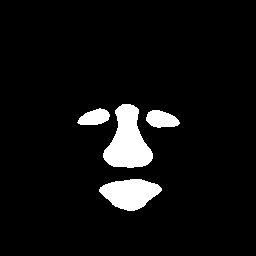} }} &
\includegraphics[width=0.12\textwidth]{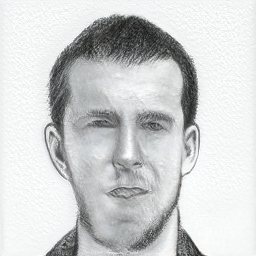} 
\makebox[0pt][r]{ \raisebox{0.0em}{\includegraphics[width=0.030\textwidth]{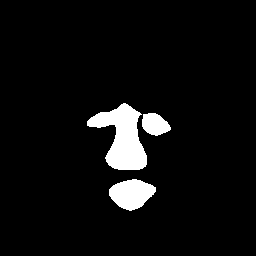} }} 
\\
\includegraphics[width=0.12\textwidth]{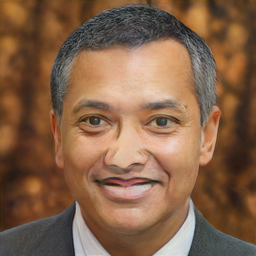} &
\includegraphics[width=0.12\textwidth]{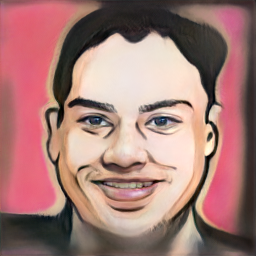} &
\includegraphics[width=0.12\textwidth]{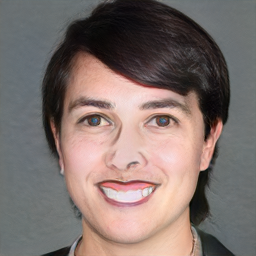} &
\includegraphics[width=0.12\textwidth]{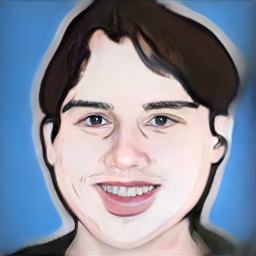} &
\includegraphics[width=0.12\textwidth]{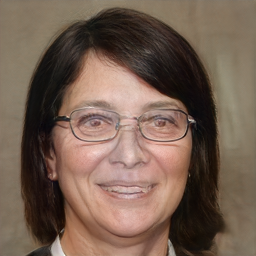} &
\includegraphics[width=0.12\textwidth]{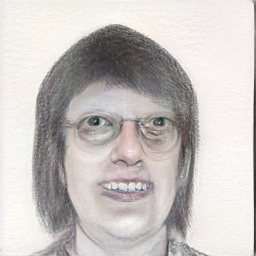} &
\includegraphics[width=0.12\textwidth]{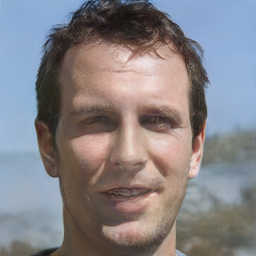} &
\includegraphics[width=0.12\textwidth]{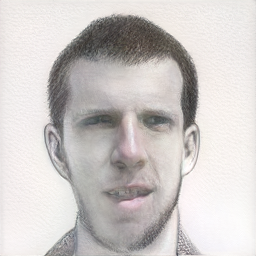} 
\\ \hline

\includegraphics[width=0.12\textwidth]{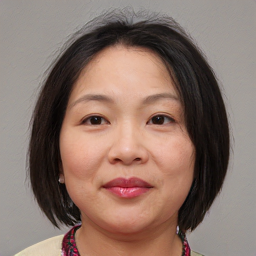} 
\makebox[0pt][r]{ \raisebox{0.0em}{\includegraphics[width=0.030\textwidth]{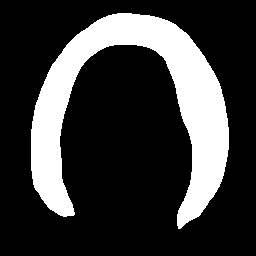} }} &
\includegraphics[width=0.12\textwidth]{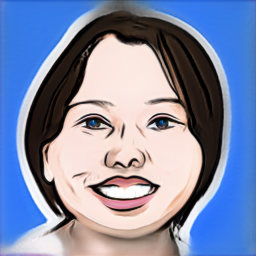} 
\makebox[0pt][r]{ \raisebox{0.0em}{\includegraphics[width=0.030\textwidth]{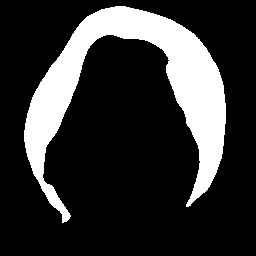} }} &
\includegraphics[width=0.12\textwidth]{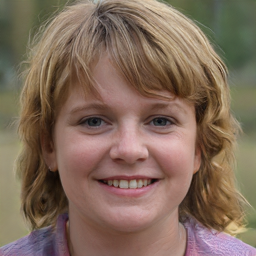} 
\makebox[0pt][r]{ \raisebox{0.0em}{\includegraphics[width=0.030\textwidth]{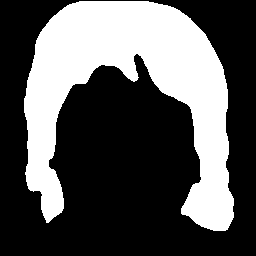} }} &
\includegraphics[width=0.12\textwidth]{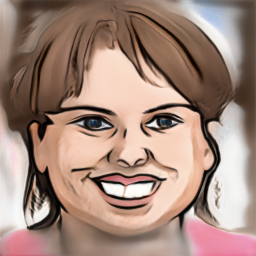} 
\makebox[0pt][r]{ \raisebox{0.0em}{\includegraphics[width=0.030\textwidth]{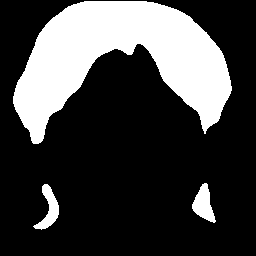} }} &
\includegraphics[width=0.12\textwidth]{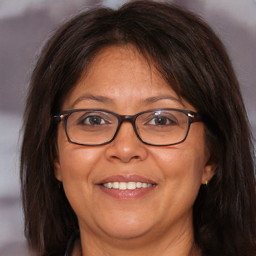} 
\makebox[0pt][r]{ \raisebox{0.0em}{\includegraphics[width=0.030\textwidth]{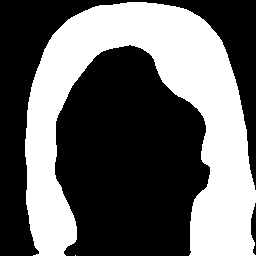} }} &
\includegraphics[width=0.12\textwidth]{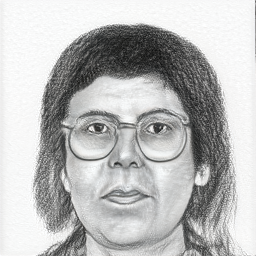} 
\makebox[0pt][r]{ \raisebox{0.0em}{\includegraphics[width=0.030\textwidth]{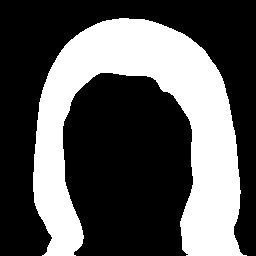} }}&
\includegraphics[width=0.12\textwidth]{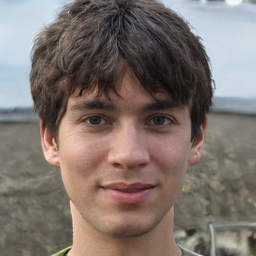} 
\makebox[0pt][r]{ \raisebox{0.0em}{\includegraphics[width=0.030\textwidth]{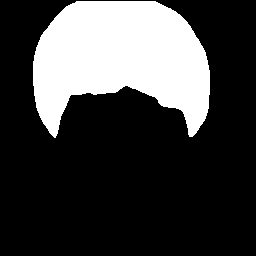} }}&
\includegraphics[width=0.12\textwidth]{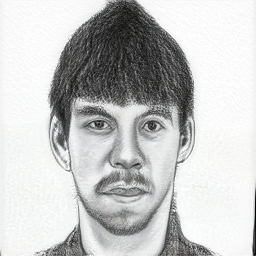} 
\makebox[0pt][r]{ \raisebox{0.0em}{\includegraphics[width=0.030\textwidth]{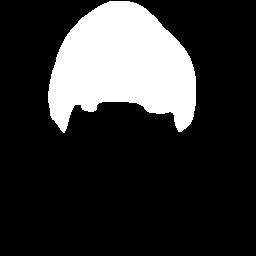} }}
\\
\includegraphics[width=0.12\textwidth]{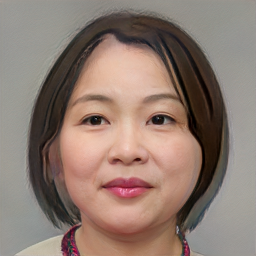} &
\includegraphics[width=0.12\textwidth]{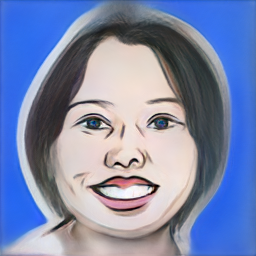} &
\includegraphics[width=0.12\textwidth]{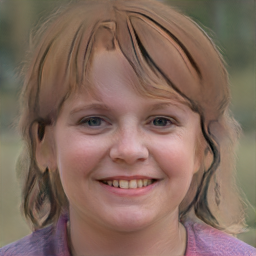} &
\includegraphics[width=0.12\textwidth]{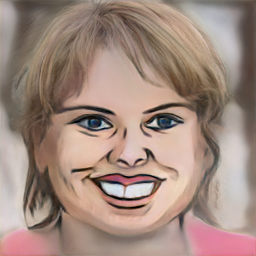} &
\includegraphics[width=0.12\textwidth]{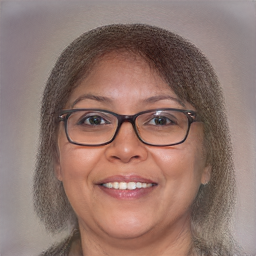} &
\includegraphics[width=0.12\textwidth]{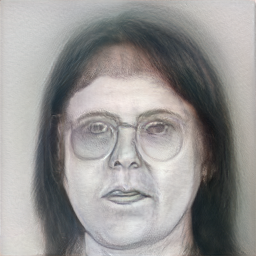} &
\includegraphics[width=0.12\textwidth]{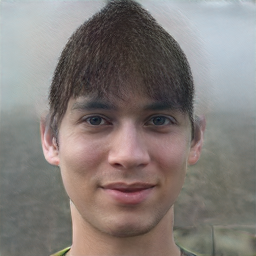} &
\includegraphics[width=0.12\textwidth]{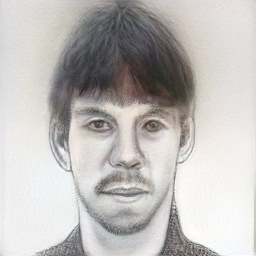} 
\end{tabular}\vspace*{-12pt}
\end{center}
\vspace*{-8pt}
\caption{DDS results on human faces. Paired images are generated from the same latent code $z^\star$. The odd columns refer to natural human face images. The second and fourth columns represent corresponding images from caricature domain. The sixth and eighth columns depict corresponding images from sketch face domain. The corresponded masks shown along with the images. The second row shows the DDS results based on the eye/nose/mouth segmentation model, while the forth row shows the DDS results  based on the hair segmentation model.  }
\label{fig:res_mouth_hair}
\end{figure*}

Our qualitative results on the face domains are shown in Fig.~\ref{fig:res_mouth_hair}. We demonstrate  natural face images and their corresponding images from either caricature domain or sketch domain along with the semantic masks. 
We present our images in pairs, where we switch the source and target domains. For example, in the first pair, DDS incorporated the eye/nose/mouth from the caricature into the natural face. Notice how the eye colour and mouth were adapted but are distinct from that of the natural image. The nose has also been copied from the caricature image, with the skin colour integrated.
In the second column, we swap the source and target domains integrating the eyes, nose and mouth from the natural image into the caricature. The natural nose is unmissable but properly integrated.

\begin{figure*}
\centering
\setlength{\tabcolsep}{0.90pt}
\begin{tabular}{c| c| c| c | c | c}
    $ Domain_{nat.}$    & 
    $\textstyle{x_{c_{car.\rightarrow nat.}}}$ &
    $\scriptstyle{DDS_{car.\rightarrow nat.}}$ &
   $ Domain_{car.}$  & 
    $\textstyle{x_{c_{nat.\rightarrow car.}}}$ &
    $\scriptstyle{DDS_{nat.\rightarrow car.}}$ 
    \\ \hline
    \includegraphics[width=0.16\textwidth]{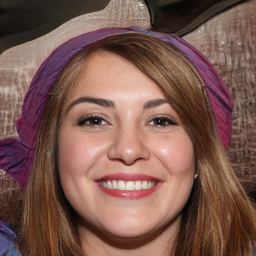} &
    \includegraphics[width=0.16\textwidth]{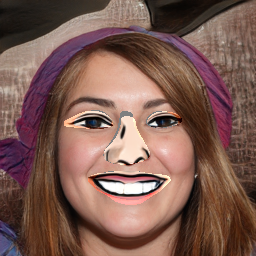} &  
    \includegraphics[width=0.16\textwidth]{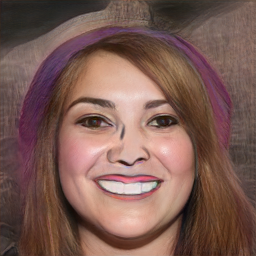} &
    \includegraphics[width=0.16\textwidth]{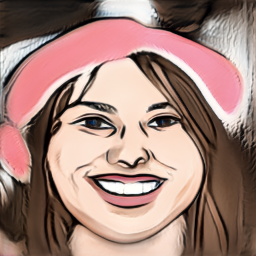} &
    \includegraphics[width=0.16\textwidth]{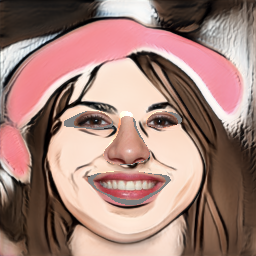} &
    \includegraphics[width=0.16\textwidth]{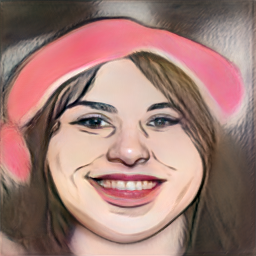}
    \end{tabular}
\vspace*{-10pt}
\caption{Comparing naive crossover to $DDS_{car. \leftrightarrow nat.}$. 
``$nat.$'' and ``$car.$'' denote the natural face and caricature domains respectively.
We show the naive crossover $x_{c_{car.\rightarrow nat.}}$ as well as our DDS results $DDS_{car.\rightarrow nat.}$. Analogous results for $nat. \rightarrow car.$ are also shown. }
\label{fig:Naive_crossover}
\end{figure*}



\noindent \textbf{Evaluation metrics:} 
We use standard metrics to evaluate our proposed technique. In each experiment, we consider the synthesised images as $\hat{x}$ and images in the domain we compare to as $x$. Note that there is no ground-truth or examples of the target dual-domain images. We thus compare the distribution of all synthesised images to three domains: source, target and the naive crossover images.
Generally, we aim for DDS images to be close to the target domain while maintaining the source features guided by the masks.
We use the following metrics in our quantitative results:
\vspace*{-6pt}
\begin{itemize}
\item Fr\'echet Inception Distance (FID)~\cite{FID} defined as $FID(\hat{x},x)~=~\left \| \mu_{\hat{x}} - \mu_{x} \right \|^2 + \Gamma (\sigma_{\hat{x}}+\sigma_{x} - 2\sqrt{\sigma_{\hat{x}} \sigma_{x}} )$, where $\mu_x$, $\sigma_x$, $\mu_{\hat{x}}$, $\sigma_{\hat{x}}$ are, respectively, the mean and covariance of $x$ and $\hat{x}$, and $\Gamma(\cdot)$ is the trace operation.

\vspace*{-6pt}
\item Structural Similarity Index Measure (SSIM)~\cite{SSIM} defined as $SSIM(\hat{x},x)~=~\frac{(2\mu_{\hat{x}}\mu_{x}+c_1)(2\sigma_{\hat{x}x}+c_2)}{(\mu_{\hat{x}}^{2}+\mu_{x}^2+c_1)(\sigma_{\hat{x}}^{2}+\sigma_{x}^2+c_2)}$,
where $c_1$ and $c_2$ stabilise the division with a weak denominator. We used $c_1=10^{-4}$ and $c_2=9\times 10^{-4}$ in all experiments as implemented in~\cite{SSIM_implementation}.

\item Peak Signal-to-Noise Ratio (PSNR)~\cite{PSNR} defined as $PSNR(\hat{x},x) =~10 \log_{10} (R^2/MSE(\hat{x},x))$, where $\scriptsize{MSE(\hat{x},x)}=~\frac{1}{n}\sum_{i=0}^{n}\left [\hat{x}_i-x_i \right ]^2$, and $R$ is the maximal in the image data. 
\end{itemize}

For each DDS experiment, we use the notation ${\mathcal{D}_s \rightarrow \mathcal{D}_t}$ to identify the source ($\mathcal{D}_s$) and target ($\mathcal{D}_t$) domains, as we integrate features from the source domain into an image from the target domain, as illustrated in Fig.~\ref{fig:Naive_crossover}.



\noindent \textbf{Quantitative results:}
Table~\ref{tab:res_FID}  presents our results.  
In each reported measure, we average results from 100 randomly sampled latent codes.
Note in particular that the scores confirm that our results in almost all the experiments have the better $FID$, $SSIM$ and $PSNR$ scores in the target domain, demonstrating the expected behaviour of the synthetic images being closer to the target domain.

\begin{table*}[]
    \centering
    \caption{Metric comparisons on the face domains.  
    $DDS_{\mathcal{D}_s\rightarrow\mathcal{D}_t}$ integrating features from the source domain into the target domain in every case. 
    $\{x_c\}$ denotes the set of all naive crossover images given by $x_{c_{\mathcal{D}_s\rightarrow\mathcal{D}_t}}$. 
    $\mathcal{M}_1$ denotes the mask of eyes/nose/mouth while $\mathcal{M}_2$ denotes hair mask.
    FID$\downarrow$, SSIM$\uparrow$ and PSNR$\uparrow$ are the metrics, where
    $\downarrow$~indicates lower is better and $\uparrow$ indicates higher is better.}
    \vspace*{-6pt}
    \label{tab:res_FID}
\setlength{\tabcolsep}{6.3pt}
\begin{tabular}{c c | c c c | c c c || c c c | c c c }  
\hline 

&  & \multicolumn{3}{c|}{ \scriptsize{$DDS_{caricature \rightarrow natural}$ } } & 
     \multicolumn{3}{c||}{ \scriptsize{$DDS_{natural\rightarrow caricature} $} } &
     \multicolumn{3}{c|}{ \scriptsize{$DDS_{sketch \rightarrow natural}$}} & 
     \multicolumn{3}{c}{ \scriptsize{$DDS_{natural \rightarrow sketch}$} } 
\\

 &  & \scriptsize{FID$\downarrow$} & \scriptsize{SSIM$\uparrow$} & \scriptsize{PSNR$\uparrow$} & 
      \scriptsize{FID$\downarrow$} & \scriptsize{SSIM$\uparrow$} & \scriptsize{PSNR$\uparrow$} & 
      \scriptsize{FID$\downarrow$} & \scriptsize{SSIM$\uparrow$} & \scriptsize{PSNR$\uparrow$} & 
      \scriptsize{FID$\downarrow$} & \scriptsize{SSIM$\uparrow$} & \scriptsize{PSNR$\uparrow$} \\  \hline

 
 \multirow{3}{*}{\small{\textit{$\mathcal{M}_1$}}}  & \small{$\mathcal{D}_s$} &  \small{281.61} & \small{0.39 } & \small{ 27.98} & \small{259.12} & \small{0.38} & \small{27.95} &  \small{195.80} & \small{0.31} & \small{27.87}  &  \small{225.31} & \small{0.33} & \small{27.81}  \\  
 
 
 & \small{$\{x_c\}$} &  \small{ 114.03} & \small{ 0.69} & \small{29.45} & \small{121.21} & \small{0.75} & \small{28.75} & \small{200.35} & \small{0.71} & \small{29.81} & \small{207.41} & \small{0.65} & \small{29.28}  \\ 
 
 
 & \small{$\mathcal{D}_t$} & 
 \small{\textbf{73.51}} & \small{\textbf{0.70}} & \small{\textbf{29.51}}  & \small{\textbf{74.51}} & \small{\textbf{0.76}} & \small{\textbf{28.76}} & \small{\textbf{79.87}} & \small{\textbf{0.73}} & \small{\textbf{29.84}} & \small{\textbf{97.65}} & \small{\textbf{0.66}} & \small{\textbf{29.29}} \\

\hline
 
 
 \multirow{3}{*}{\small{\textit{$\mathcal{M}_2$}}}  & \small{$\mathcal{D}_s$} &  \small{296.61} & \small{0.47 } & \small{ 27.99} & \small{290.63} & \small{0.42} & \small{27.92} & \small{245.61} & \small{0.38} & \small{27.87}  & \small{262.90} & \small{0.36} & \small{27.83}  \\ 
 

 & \small{$\{x_c\}$} &  \small{ 155.91} & \small{0.73 } & \small{ 29.63} & \small{95.30} & \small{0.71} & \small{28.43} &  \small{177.03} & \small{\textbf{0.66}} & \small{29.25}  & \small{152.41} & \small{0.61} & \small{\textbf{28.54}}  \\ 

 
 & \small{$\mathcal{D}_t$} & \small{\textbf{ 87.28 }} & \small{\textbf{  0.74}} & \small{\textbf{ 29.65}} & \small{\textbf{76.96}} & \small{\textbf{0.72}} & \small{\textbf{28.45}} & \small{\textbf{90.58}} & \small{0.64} & \small{\textbf{29.31}} & \small{\textbf{94.83}} & \small{\textbf{0.62}} & \small{28.49} \\

\hline

 
\end{tabular}
\end{table*}

\begin{figure}
    \centering
    \includegraphics[width=0.65\linewidth]{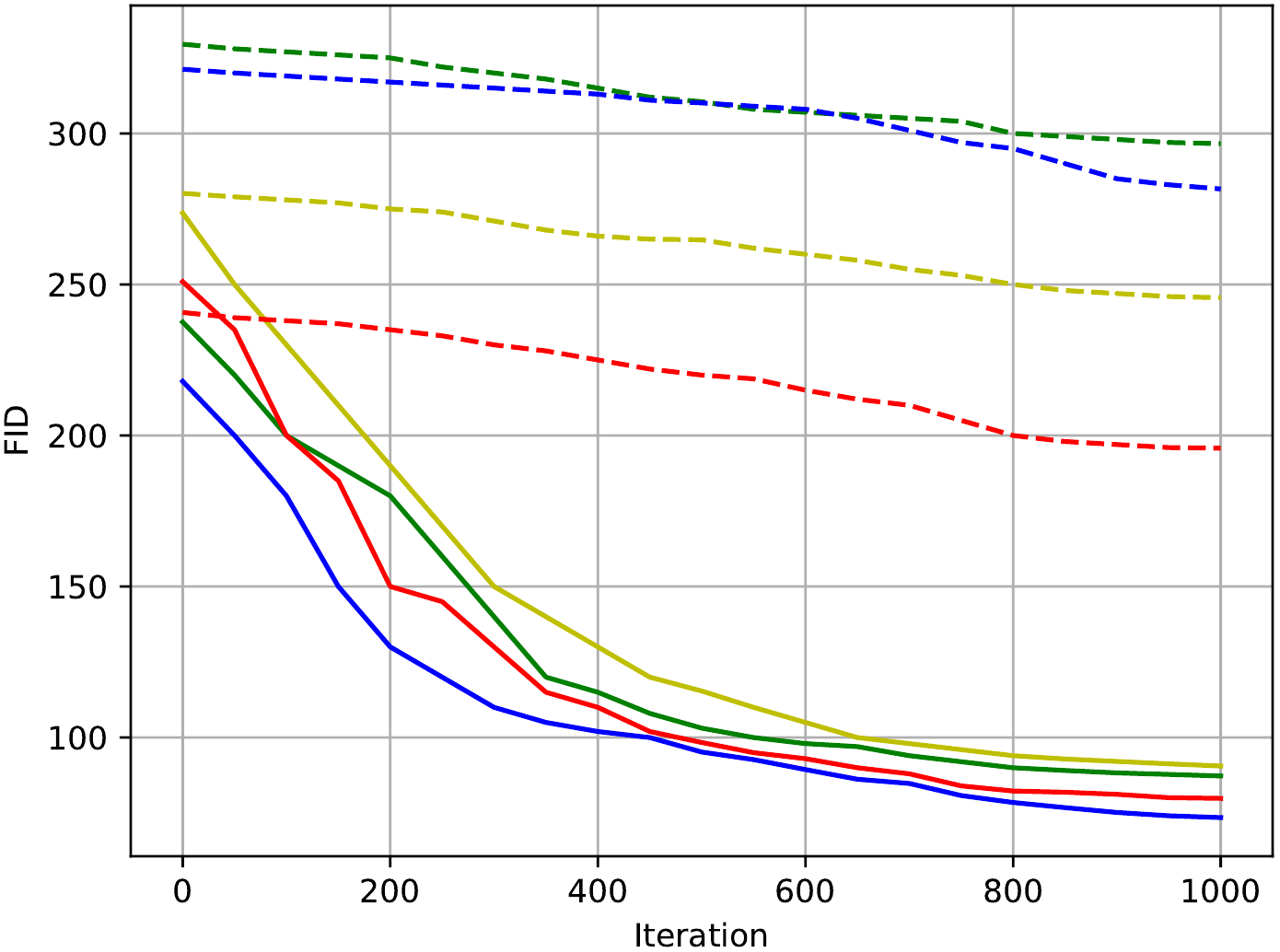} 
    \includegraphics[width=0.3\linewidth]{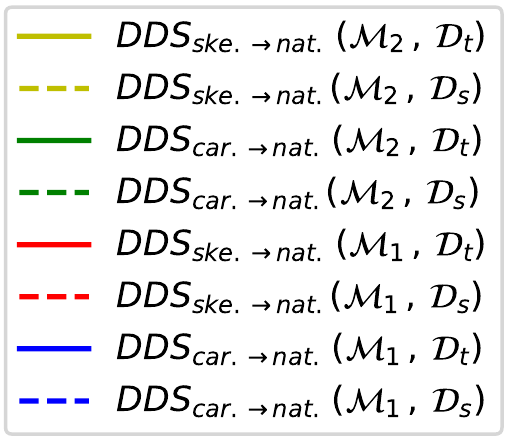}
    \vspace*{-10pt}
    \caption{FID during latent code optimisation iterations, comparing the dual-domain images to source (dashed) and target (solid) domains for 4 experiments. Dual-domain images get closer to both domains during optimisation but significantly closer to target.}
    \label{fig:FID_iterations}
\end{figure}

\noindent \textbf{Monitoring latent code optimisation:} Fig.~\ref{fig:FID_iterations} plots the FID metric at each iteration of the latent optimising process. We consider $DDS_{car. \rightarrow nat.}$ and $DDS_{sketch \rightarrow nat.}$, thus using the caricature and sketch images as source and the natural faces as target with the two segmentation models $\mathcal{M}_1$ and $\mathcal{M}_2$. We observe FID during optimisation in each case. Fig.~\ref{fig:FID_iterations} showcases how the dual-domain synthesised image gets closer to both domains but importantly significantly, and more quickly, closer to the target domain, and remains distinct from either domain at the end of the optimisation.

\begin{figure}
    \centering
    \includegraphics[width=0.8\linewidth]{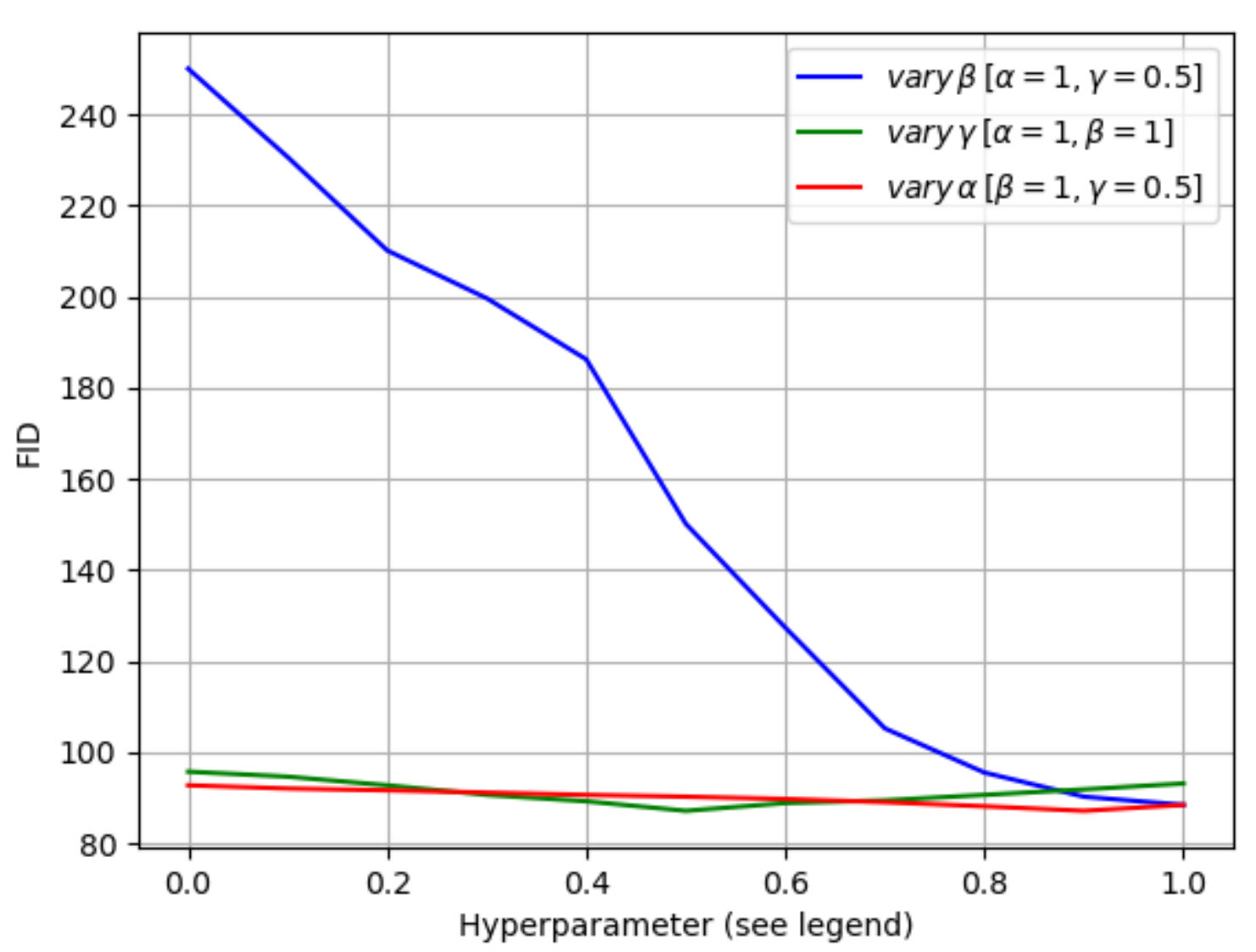} 
    \vspace*{-10pt}
    \caption{Impact of $\alpha$, $\beta$ and $\gamma$ in loss function (Eq.~\ref{eq:loss}) on natural face (target) and caricature domains (source) with the hair mask.  FID values are computed over the target domain ($\mathcal{D}_t$).}
    \label{fig:hyperparameters}
\end{figure}

\noindent \textbf{Robustness to hyperparameters:} We next assess the robustness of our results to the parametric weights of the loss function in Eq.~\ref{eq:loss}. We performed this experiment on the domains of natural human faces and caricature, and the hair segmentation model, and we consider FID over the target domain ($\mathcal{D}_t$) as a metric to evaluate the results of several experiments. Fig.~\ref{fig:hyperparameters} demonstrates that by considering $ {\alpha , \beta , \gamma } $ as ${0.9,1,0.5}$ respectively we can have the best FID in this experiment. It can be perceived from Fig.~\ref{fig:hyperparameters} that $\beta$ is the most sensitive hyperparameter in Eq.~\ref{eq:loss}. The reason for this sensitivity is that $\beta$ weights the target loss ($\mathcal{L}_t$) and these FID values are computed over the target domain. Therefore, when $\beta$ is equal to zero, we do not consider the target loss ($\mathcal{L}_t$) in our optimisation. 

\begin{figure}[t]
\centering
\begin{center}
\setlength{\tabcolsep}{1.0pt}
\begin{tabular}{ c c | c c } 

\includegraphics[width=0.11\textwidth]{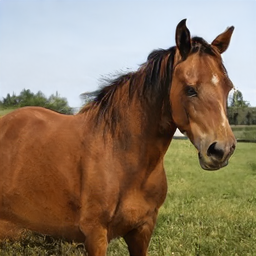}
 \makebox[0pt][r]{ \raisebox{0.0em}{\includegraphics[width=0.030\textwidth]{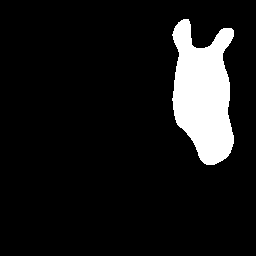} }} &
\includegraphics[width=0.11\textwidth]{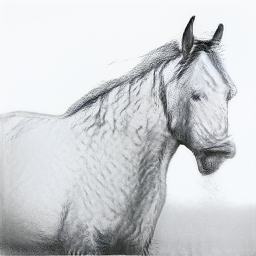}
 \makebox[0pt][r]{ \raisebox{0.0em}{\includegraphics[width=0.030\textwidth]{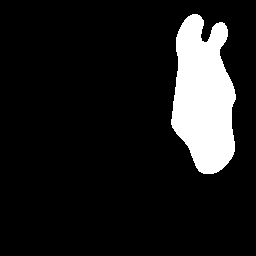} }} &
\includegraphics[width=0.11\textwidth]{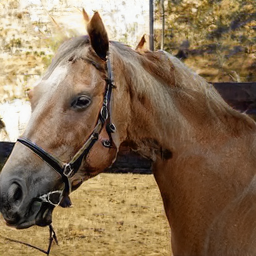}
 \makebox[0pt][r]{ \raisebox{0.0em}{\includegraphics[width=0.030\textwidth]{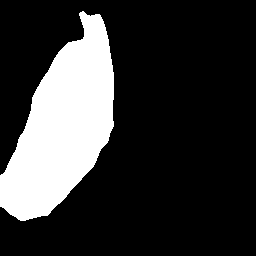} }} &
\includegraphics[width=0.11\textwidth]{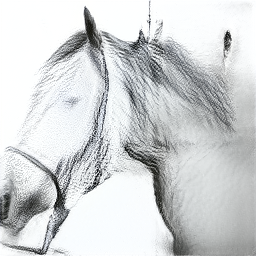}
 \makebox[0pt][r]{ \raisebox{0.0em}{\includegraphics[width=0.030\textwidth]{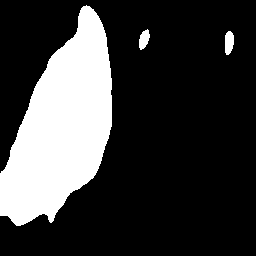} }}
\\
\includegraphics[width=0.11\textwidth]{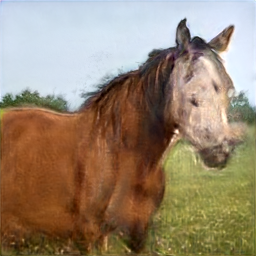}&
\includegraphics[width=0.11\textwidth]{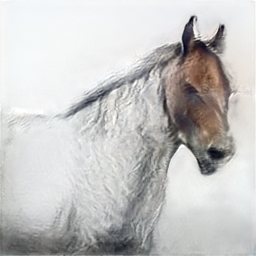}&
\includegraphics[width=0.11\textwidth]{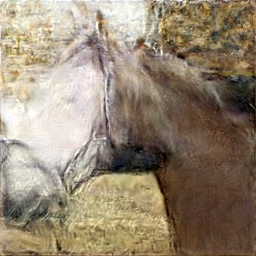}&
\includegraphics[width=0.11\textwidth]{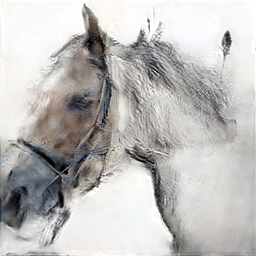}
\\ 

\end{tabular}
\end{center}
\vspace*{-23pt}
\caption{DDS results on horse domains. Odd columns are from natural horse domain, and even columns are from sketch horse domain. Segmentation masks are incorporated.  
Second row shows the corresponding dual-domain images. }
\label{fig:horse_res}
\end{figure}

\begin{figure}
\centering
\begin{center}
\setlength{\tabcolsep}{0.6pt}
\begin{tabular}{ c c | c c }

 \includegraphics[width=0.11\textwidth]{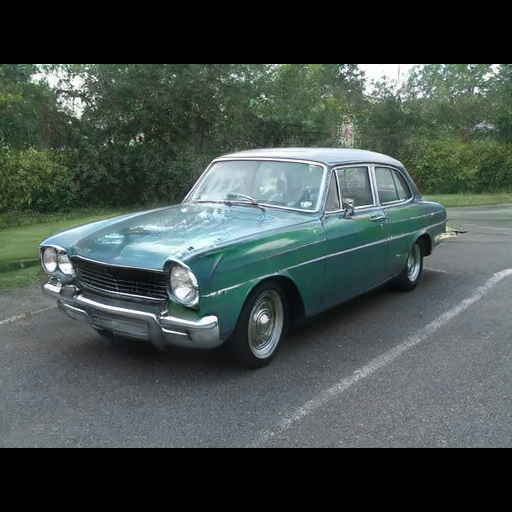}
 \makebox[0pt][r]{ \raisebox{0.5em}{\includegraphics[width=0.030\textwidth]{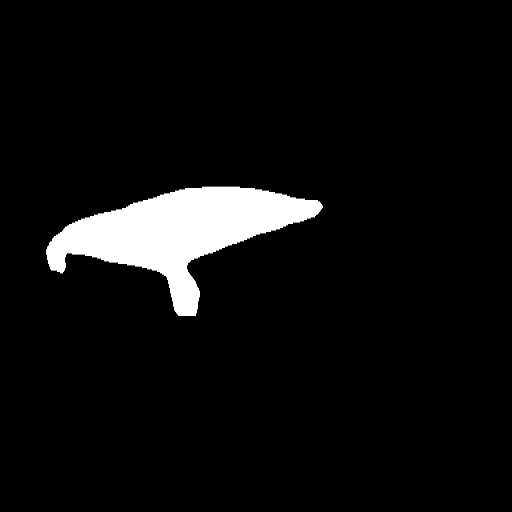} }} &
\includegraphics[width=0.11\textwidth]{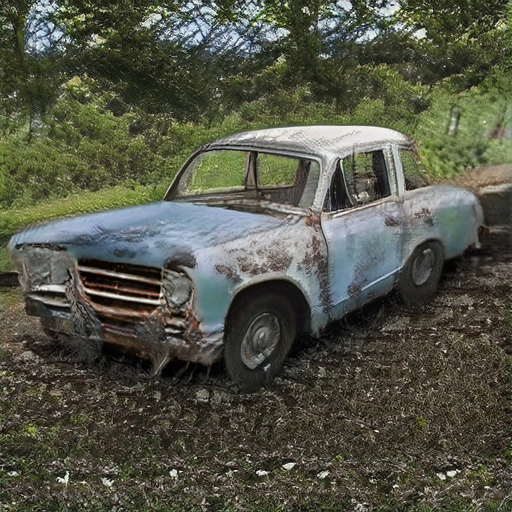}
 \makebox[0pt][r]{ \raisebox{0.0em}{\includegraphics[width=0.030\textwidth]{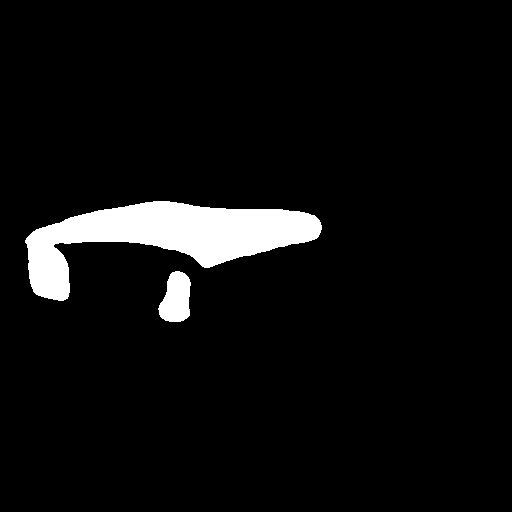} }}&
\includegraphics[width=0.11\textwidth]{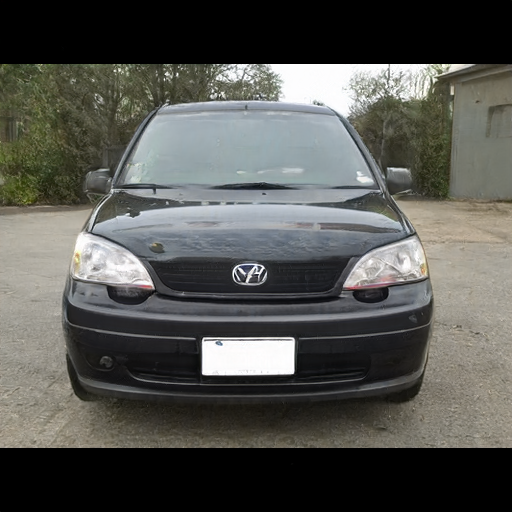}
 \makebox[0pt][r]{ \raisebox{0.5em}{\includegraphics[width=0.030\textwidth]{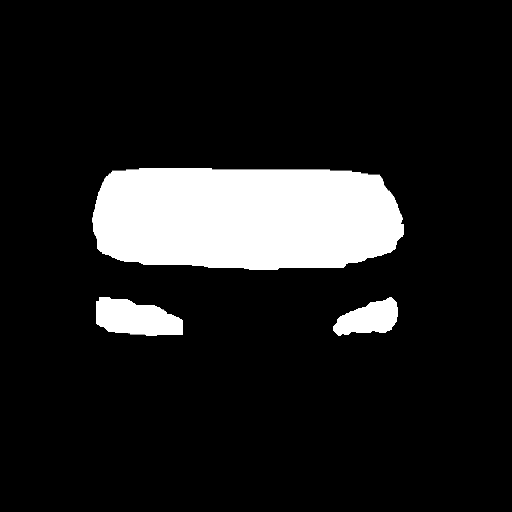} }} &
\includegraphics[width=0.11\textwidth]{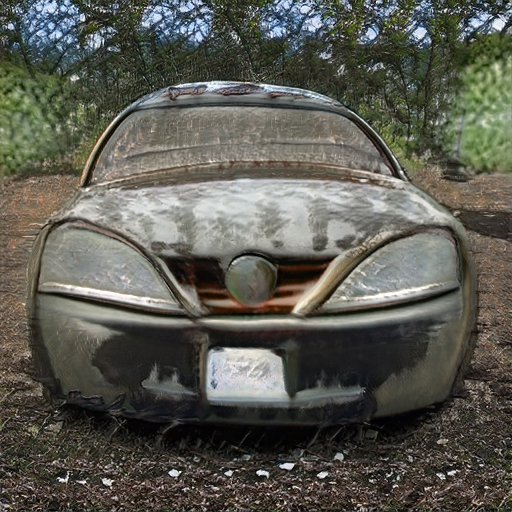}
 \makebox[0pt][r]{ \raisebox{0.0em}{\includegraphics[width=0.030\textwidth]{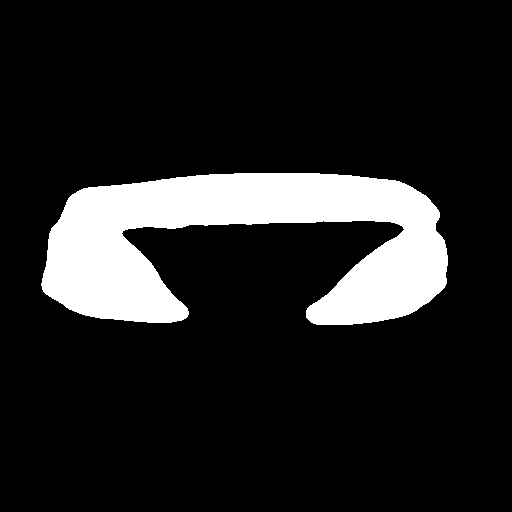} }}

\\
\includegraphics[width=0.11\textwidth]{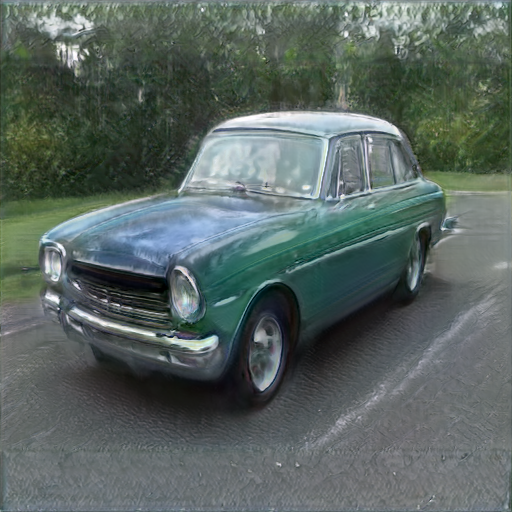}&
\includegraphics[width=0.11\textwidth]{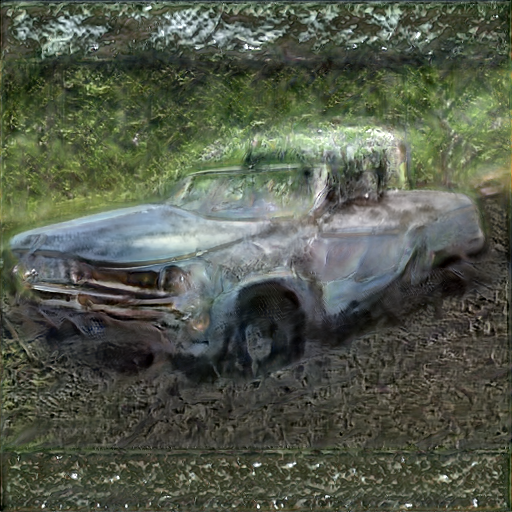}&
\includegraphics[width=0.11\textwidth]{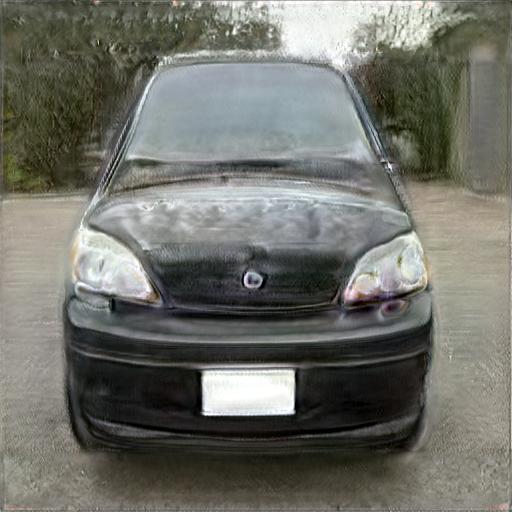}&
\includegraphics[width=0.11\textwidth]{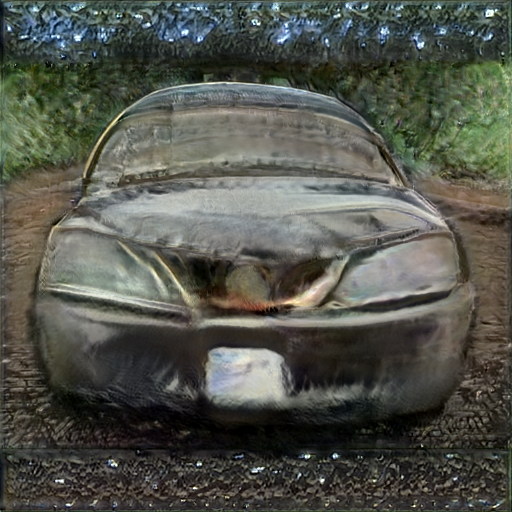}
\\
 
\end{tabular}
\end{center}
\vspace*{-23pt}
\caption{DDS results on car domains. Odd columns are standard car images, and even columns are abandoned cars. Segmentation masks are incorporated. 
Second row shows the DDS results.}
\label{fig:car_res}
\end{figure}

\begin{figure*}[t!]
\centering
\begin{center}
\setlength{\tabcolsep}{0.01pt}
\begin{tabular}{ c c | c c | c c | c c}
\includegraphics[width=0.12\textwidth]{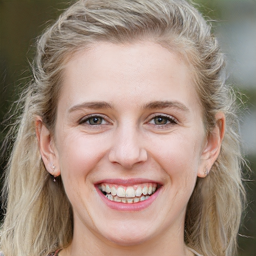} 
\makebox[0pt][r]{ \raisebox{0em}{\includegraphics[width=0.030\textwidth]{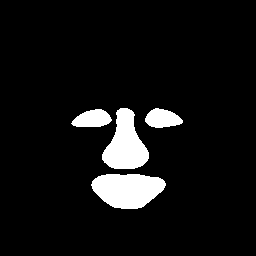} }}&
\includegraphics[width=0.12\textwidth]{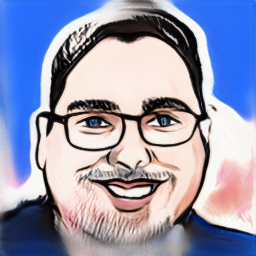} 
\makebox[0pt][r]{ \raisebox{0em}{\includegraphics[width=0.030\textwidth]{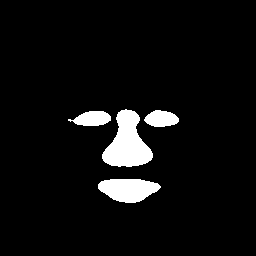} }} &


\includegraphics[width=0.12\textwidth]{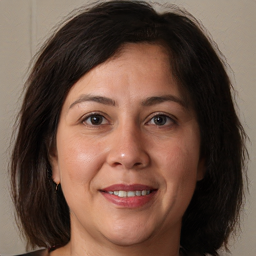} 
\makebox[0pt][r]{ \raisebox{0em}{\includegraphics[width=0.030\textwidth]{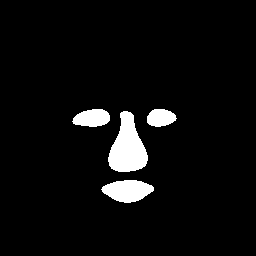} }}&
\includegraphics[width=0.12\textwidth]{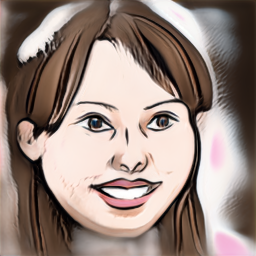} 
\makebox[0pt][r]{ \raisebox{0em}{\includegraphics[width=0.030\textwidth]{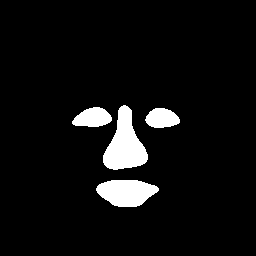} }}&
\includegraphics[width=0.12\textwidth]{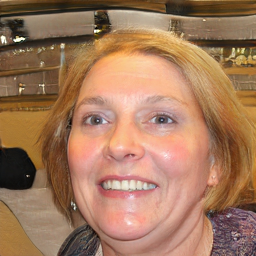} 
\makebox[0pt][r]{ \raisebox{0em}{\includegraphics[width=0.030\textwidth]{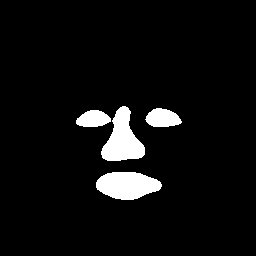} }}&
\includegraphics[width=0.12\textwidth]{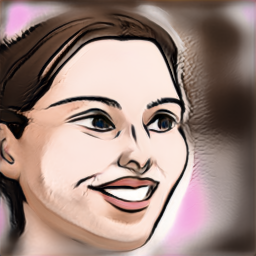} 
\makebox[0pt][r]{ \raisebox{0em}{\includegraphics[width=0.030\textwidth]{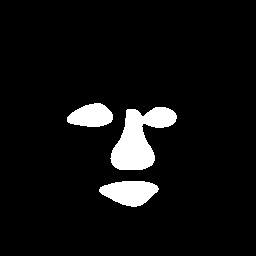} }} &
\includegraphics[width=0.12\textwidth]{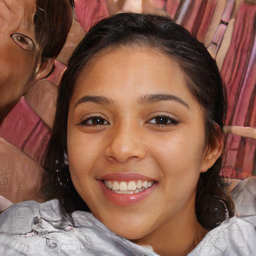} 
\makebox[0pt][r]{ \raisebox{0em}{\includegraphics[width=0.030\textwidth]{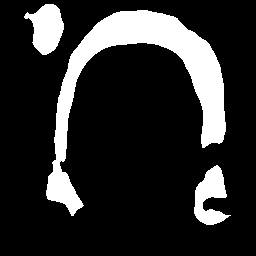} }}&
\includegraphics[width=0.12\textwidth]{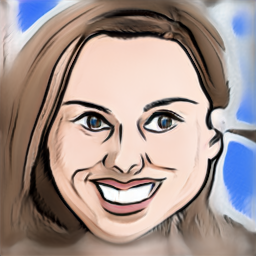} 
\makebox[0pt][r]{ \raisebox{0em}{\includegraphics[width=0.030\textwidth]{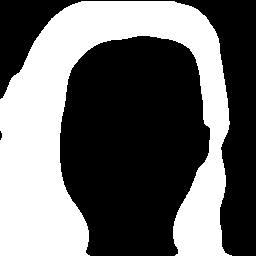} }} 

\\
\includegraphics[width=0.12\textwidth]{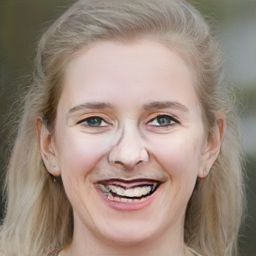} &
\includegraphics[width=0.12\textwidth]{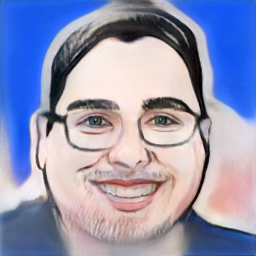} &
\includegraphics[width=0.12\textwidth]{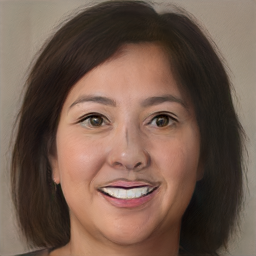} &
\includegraphics[width=0.12\textwidth]{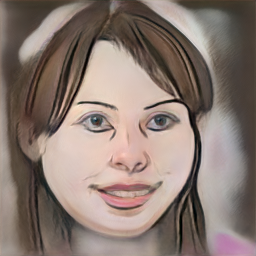} &
\includegraphics[width=0.12\textwidth]{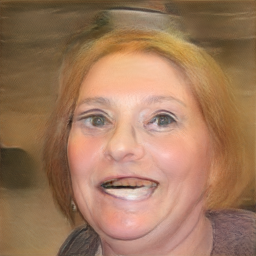} &
\includegraphics[width=0.12\textwidth]{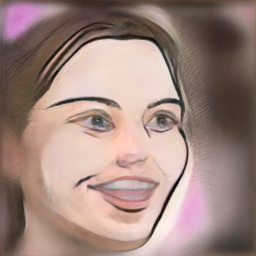} &
\includegraphics[width=0.12\textwidth]{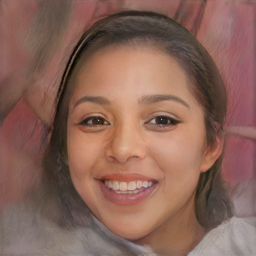} &
\includegraphics[width=0.12\textwidth]{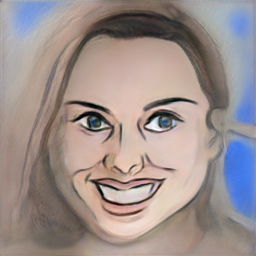} 
\end{tabular}
\end{center}
\vspace*{-23pt}
\caption{The DDS results on images from unpaired latent codes ($z^\star_{1}, z^\star_{2}$). 
We show the segmentation mask in each case. 
Successful examples (col 1-4) show that DDS can bridge small pose differences. 
Failures (col 5-8) are due to larger pose variations.}
\label{fig:unpaired}
\end{figure*}

\begin{figure}[!h]
\centering
\begin{center}
\setlength{\tabcolsep}{0.09pt}
\begin{tabular}{ c c | c c } 
 \includegraphics[width=0.11\textwidth]{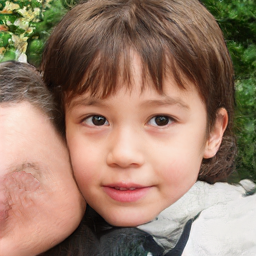}
 \makebox[0pt][r]{ \raisebox{0.0em}{\includegraphics[width=0.028\textwidth]{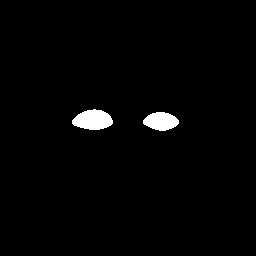} }} &
 \includegraphics[width=0.11\textwidth]{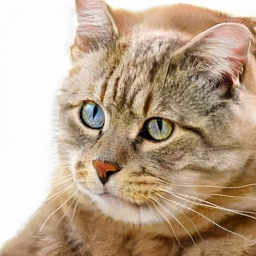}
 \makebox[0pt][r]{ \raisebox{0.0em}{\includegraphics[width=0.028\textwidth]{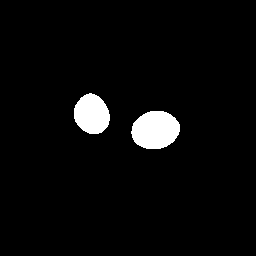} }} &
  \includegraphics[width=0.11\textwidth]{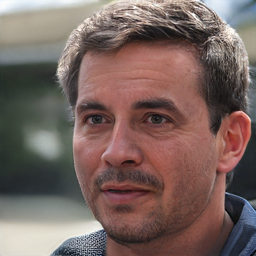}
 \makebox[0pt][r]{ \raisebox{0.0em}{\includegraphics[width=0.028\textwidth]{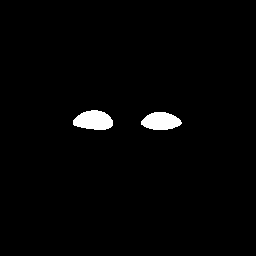} }} &
 \includegraphics[width=0.11\textwidth]{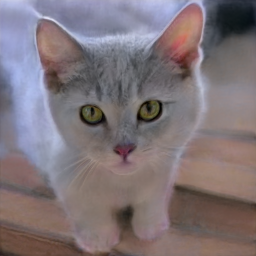}
\makebox[0pt][r]{ \raisebox{0.0em}{\includegraphics[width=0.028\textwidth]{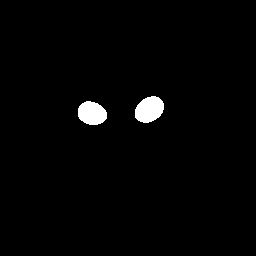} }} 
 \\
\includegraphics[width=0.11\textwidth]{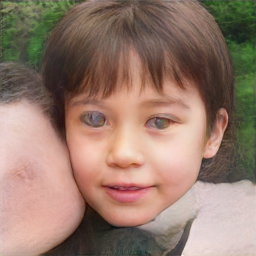}&
\includegraphics[width=0.11\textwidth]{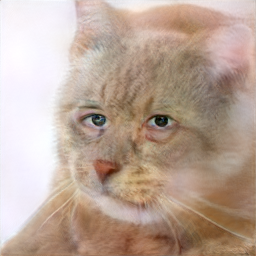}&
\includegraphics[width=0.11\textwidth]{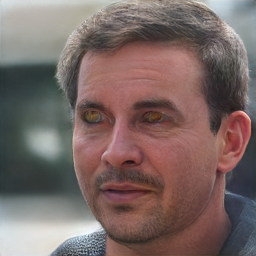}&
\includegraphics[width=0.11\textwidth]{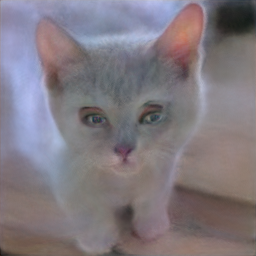}
\end{tabular}
\end{center}
\vspace*{-23pt}
\caption{DDS results on unpaired human and cat images based on the segmentation model on eyes. First row shows the unpaired images with their segmentation mask. Second row shows human faces with the cat eyes and cats with the human eyes.}
\label{fig:human_cat_eyes}
\vspace*{-12pt}
\end{figure}

\subsection{DDS on Horse and Car Domains} 

We performed further experiments on horse and car images. We used the StyleGAN2~\cite{stylegan2} model to generate natural horse images from~\cite{StyleGAN2_models} which is trained on LSUN Horses~\cite{LSUN}. 
We collected 10 sketch horses for the purpose of this experiment through a web search.
We scale these images to $256 \times 256$ and use these 10 images for the few-shotGAN far-domain adaptation~\cite{fewShotGAN}.
We train our one-shot segmentation model to segment the head of the horse. 
Fig.~\ref{fig:horse_res} shows qualitative results. 



For car images, we used the StyleGAN2~\cite{stylegan2} model to generate the standard car images from~\cite{StyleGAN2_models}, which is trained on LSUN Cars~\cite{LSUN}, 
and used 10 abandoned car images from the few-shotGAN implementation~\cite{fewshotGAN_github} for few-shot adaptation. 
Fig.~\ref{fig:car_res} shows the qualitative results. The one-shot segmentation model is trained to segment the bonnet and lights of a car. 
Note that for both cars and horses, we use the $conv$ layers of AlexNet~\cite{AlexNet} for the perceptual loss. This performs better than VGG~\cite{VGG} (see appendix).


\subsection {Unpaired latents, different objects, limitations} 

Failures in the domain adaptation (Sec.~\ref{sec:fewshot}) or in the segmentation masks (Sec.~\ref{sec:repurposingGAN}) will negatively impact DDS.
In all previous results, we pass the same latent code $z^\star$ to the two generative models. Thanks to the adaptation algorithm, this guarantees the same pose in the corresponding images.

In Fig.~\ref{fig:unpaired} , we show that DDS can work with unpaired latent codes with small pose variations.
In all the domains and examples, DDS assumes the corresponding source and target images $x_s, x_t$ are roughly aligned in pose.
 Fig.~\ref{fig:unpaired} also shows that DDS cannot recover from large pose variations in the unpaired images where the segmentation masks are not aligned.

We performed further unpaired experiments on cat and natural face images. We used the StyleGAN2~\cite{stylegan2} model to generate cat images from~\cite{StyleGAN2_models} which is trained on LSUN Cats~\cite{LSUN}. We used a segmentation model on eyes with the purpose of generating dual-domain images of human face with the cat eyes and vice versa. Fig.~\ref{fig:human_cat_eyes} shows successful dual-domain synthesis.
This demonstrates that DDS is not limited to examples when one object types, and can be used with two different many-shot styleGANs and are not adapted from one another, when masks are aligned.
We present further results on this in appendix where both the source and target domains are from the few-shotGAN. 
The results in Fig.~\ref{fig:human_cat_eyes} demonstrate the potential of DDS on two unpaired examples from unrelated GANs and distinct segmentation models. 
However, successful examples are challenging to find due to the variation in pose when randomly sampling latent variables. 
This forms a limitation that we wish to address in future work, by utilising segmentation alignment approaches. 

We also noted that in contrast with prior assumption~\cite{unreasonable_perceptual_metric}, the choice of backbone for the perceptual loss impacts DDS results and is domain specific.
We present further results on this in appendix. 
Further exploration is needed to ensure robustness to backbone choice in the perceptual loss.

\section{Conclusion}

We presented Dual-Domain Synthesis framework, an approach to generate visually convincing images that contain features from different domains.
Our generated images are obtained by combining two correspondence images guided by part-based segmentation from a one-shot segmentation model.
We demonstrated the approach on several image classes (face, car, horse, cat), multiple domains (natural, sketch, caricature, abandoned) and various part-based segmentation models (eye, nose, mouth, hair, car bonnet). 

Extensive experiments illustrate the potential of the approach, both in terms of qualitative and quantitative assessment. While the majority of the results focus on paired latent codes and object (e.g. faces), the approach can be extended to distinct GANs trained independently and unpaired latent codes, assuming part-based alignment.

{\small
\bibliographystyle{ieee_fullname}
\bibliography{NTIRE_workshop}

\begin{thebibliography}{10}\itemsep=-1pt

\bibitem{SSIM_implementation}
{SSIM Implementation}.
\newblock "\url{https://github.com/Po-Hsun-Su/pytorch-ssim/}", 2017.
\newblock [Online].

\bibitem{StyleGAN2_models}
{StyleGAN2 Models}.
\newblock "\url{https://nvlabs-fi-cdn.nvidia.com/stylegan2/networks/}", 2019.
\newblock [Online].

\bibitem{stylegan_github}
{PyTorch Implementation of Analyzing and Improving the Image Quality of
  StyleGAN}.
\newblock "\url{https://github.com/rosinality/stylegan2-pytorch/}", 2020.
\newblock [Online].

\bibitem{fewshotGAN_github}
{Implementation of Few-shot Image Generation via Cross-domain Correspondence}.
\newblock "\url{https://github.com/utkarshojha/few-shot-gan-adaptation/}",
  2021.
\newblock [Online].

\bibitem{repurposeGAN_github}
{Implementation of Repurposing GANs for One-shot Semantic Part Segmentation}.
\newblock "\url{https://github.com/bryandlee/repurpose-gan/}", 2021.
\newblock [Online].

\bibitem{image2stylegan}
Rameen Abdal, Yipeng Qin, and Peter Wonka.
\newblock {Image2StyleGAN: How to Embed Images into the StyleGAN Latent Space?}
\newblock In {\em Proceedings of the IEEE/CVF International Conference on
  Computer Vision}, pages 4432--4441, 2019.

\bibitem{image2stylegan++}
Rameen Abdal, Yipeng Qin, and Peter Wonka.
\newblock {Image2StyleGAN++: How to Edit the Embedded Images?}
\newblock In {\em Proceedings of the IEEE/CVF Conference on Computer Vision and
  Pattern Recognition}, pages 8296--8305, 2020.

\bibitem{text_segmentation}
Mohammed Al-Rawi, Dena Bazazian, and Ernest Valveny.
\newblock {Can Generative Adversarial Networks Teach Themselves Text
  Segmentation?}
\newblock In {\em Proceedings of the IEEE/CVF International Conference on
  Computer Vision Workshops}, 2019.

\bibitem{LatentStyleEdit}
Edo Collins, Raja Bala, Bob Price, and Sabine Susstrunk.
\newblock {Editing in Style: Uncovering the Local Semantics of GANs}.
\newblock In {\em Proceedings of the IEEE/CVF Conference on Computer Vision and
  Pattern Recognition}, pages 5771--5780, 2020.

\bibitem{cycleGAN_segmentation}
Wolfgang Fuhl, David Geisler, Wolfgang Rosenstiel, and Enkelejda Kasneci.
\newblock {The Applicability of Cycle {GANs} for Pupil and Eyelid Segmentation,
  Data Generation and Image Refinement}.
\newblock In {\em Proceedings of the IEEE/CVF International Conference on
  Computer Vision Workshops}, 2019.

\bibitem{GAN}
Ian Goodfellow, Jean Pouget-Abadie, Mehdi Mirza, Bing Xu, David Warde-Farley,
  Sherjil Ozair, Aaron Courville, and Yoshua Bengio.
\newblock {Generative Adversarial Nets}.
\newblock {\em Advances in Neural Information Processing Systems}, 27, 2014.

\bibitem{portrait_maskguided}
Shuyang Gu, Jianmin Bao, Hao Yang, Dong Chen, Fang Wen, and Lu Yuan.
\newblock {Mask-guided Portrait Editing with Conditional {GANs}}.
\newblock In {\em Proceedings of the IEEE/CVF Conference on Computer Vision and
  Pattern Recognition}, pages 3436--3445, 2019.

\bibitem{ResNet}
Kaiming He, Xiangyu Zhang, Shaoqing Ren, and Jian Sun.
\newblock {Deep Residual Learning for Image Recognition}.
\newblock In {\em Proceedings of the IEEE Conference on Computer Vision and
  Pattern Recognition}, pages 770--778, 2016.

\bibitem{FID}
Martin Heusel, Hubert Ramsauer, Thomas Unterthiner, Bernhard Nessler, and Sepp
  Hochreiter.
\newblock {GANs Trained by a Two Time-Scale Update Rule Converge to a Local
  Nash Equilibrium}.
\newblock {\em Advances in Neural Information Processing Systems}, 30, 2017.

\bibitem{PSNR}
Quan Huynh-Thu and Mohammed Ghanbari.
\newblock {Scope of Validity of {PSNR} in Image/Video Quality Assessment}.
\newblock {\em Electronics Letters}, 44(13):800--801, 2008.

\bibitem{Squeezenet}
Forrest~N Iandola, Song Han, Matthew~W Moskewicz, Khalid Ashraf, William~J
  Dally, and Kurt Keutzer.
\newblock {SqueezeNetX: {AlexNet-level} Accuracy with 50x Fewer Parameters and
  {$<$} 0.5MB Model Size}.
\newblock {\em arXiv preprint arXiv:1602.07360}, 2016.

\bibitem{perceptual_loss}
Justin Johnson, Alexandre Alahi, and Li Fei-Fei.
\newblock {Perceptual Losses for Real-Time Style Transfer and
  Super-Resolution}.
\newblock In {\em European Conference on Computer Vision}, pages 694--711.
  Springer, 2016.

\bibitem{stylegan1}
Tero Karras, Samuli Laine, and Timo Aila.
\newblock {A Style-Based Generator Architecture for Generative Adversarial
  Networks}.
\newblock In {\em Proceedings of the IEEE/CVF Conference on Computer Vision and
  Pattern Recognition}, pages 4401--4410, 2019.

\bibitem{stylegan2}
Tero Karras, Samuli Laine, Miika Aittala, Janne Hellsten, Jaakko Lehtinen, and
  Timo Aila.
\newblock {Analyzing and Improving the Image Quality of StyleGAN}.
\newblock In {\em Proceedings of the IEEE/CVF Conference on Computer Vision and
  Pattern Recognition}, pages 8110--8119, 2020.

\bibitem{adam}
Diederik~P Kingma and Jimmy Ba.
\newblock {Adam: A Method for Stochastic Optimization}.
\newblock {\em arXiv preprint arXiv:1412.6980}, 2014.

\bibitem{AlexNet}
Alex Krizhevsky.
\newblock {One Weird Trick for Parallelizing Convolutional Neural Networks}.
\newblock {\em arXiv preprint arXiv:1404.5997}, 2014.

\bibitem{imageInpaintingIrregular}
Guilin Liu, Fitsum~A Reda, Kevin~J Shih, Ting-Chun Wang, Andrew Tao, and Bryan
  Catanzaro.
\newblock {Image Inpainting for Irregular Holes using Partial Convolutions}.
\newblock In {\em Proceedings of the European Conference on Computer Vision},
  pages 85--100, 2018.

\bibitem{PDGAN}
Hongyu Liu, Ziyu Wan, Wei Huang, Yibing Song, Xintong Han, and Jing Liao.
\newblock {PD-GAN}: {Probabilistic Diverse {GAN} for Image Inpainting}.
\newblock In {\em Proceedings of the IEEE/CVF Conference on Computer Vision and
  Pattern Recognition}, pages 9371--9381, 2021.

\bibitem{image_translation}
Ming-Yu Liu, Thomas Breuel, and Jan Kautz.
\newblock {Unsupervised Image-to-Image Translation Networks}.
\newblock In {\em Advances in Neural Information Processing Systems}, pages
  700--708, 2017.

\bibitem{fewShotGAN}
Utkarsh Ojha, Yijun Li, Jingwan Lu, Alexei~A Efros, Yong~Jae Lee, Eli
  Shechtman, and Richard Zhang.
\newblock {Few-shot Image Generation via Cross-domain Correspondence}.
\newblock In {\em Proceedings of the IEEE/CVF Conference on Computer Vision and
  Pattern Recognition}, pages 10743--10752, 2021.

\bibitem{SPADE}
Taesung Park, Ming-Yu Liu, Ting-Chun Wang, and Jun-Yan Zhu.
\newblock {Semantic Image Synthesis with Spatially-Adaptive Normalization}.
\newblock In {\em Proceedings of the IEEE/CVF Conference on Computer Vision and
  Pattern Recognition}, pages 2337--2346, 2019.

\bibitem{few-shot-adaptation}
Esther Robb, Wen-Sheng Chu, Abhishek Kumar, and Jia-Bin Huang.
\newblock {Few-shot Adaptation of Generative Adversarial Networks}.
\newblock {\em arXiv preprint arXiv:2010.11943}, 2020.

\bibitem{forground_segmentation}
Dimitrios Sakkos, Edmond~SL Ho, and Hubert~PH Shum.
\newblock {Illumination-aware Multi-Task {GANs} for Foreground Segmentation}.
\newblock {\em IEEE Access}, 7:10976--10986, 2019.

\bibitem{VGG}
Karen Simonyan and Andrew Zisserman.
\newblock {Very Deep Convolutional Networks for Large-Scale Image Recognition}.
\newblock {\em arXiv preprint arXiv:1409.1556}, 2014.

\bibitem{sofiiuk2021foreground}
Konstantin Sofiiuk, Polina Popenova, and Anton Konushin.
\newblock Foreground-aware semantic representations for image harmonization.
\newblock In {\em Proceedings of the IEEE/CVF Winter Conference on Applications
  of Computer Vision}, pages 1620--1629, 2021.

\bibitem{repurposingGAN}
Nontawat Tritrong, Pitchaporn Rewatbowornwong, and Supasorn Suwajanakorn.
\newblock {Repurposing {GANs} for One-shot Semantic Part Segmentation}.
\newblock In {\em Proceedings of the IEEE/CVF Conference on Computer Vision and
  Pattern Recognition}, pages 4475--4485, 2021.

\bibitem{sketch_face}
Xiaogang Wang and Xiaoou Tang.
\newblock {Face Photo-Sketch Synthesis and Recognition}.
\newblock {\em IEEE Transactions on Pattern Analysis and Machine Intelligence},
  31(11):1955--1967, 2008.

\bibitem{SSIM}
Zhou Wang, Alan~C Bovik, Hamid~R Sheikh, and Eero~P Simoncelli.
\newblock {Image Quality Assessment: From Error Visibility to Structural
  Similarity}.
\newblock {\em IEEE Transactions on Image Processing}, 13(4):600--612, 2004.

\bibitem{wu2019gp}
Huikai Wu, Shuai Zheng, Junge Zhang, and Kaiqi Huang.
\newblock {GP-GAN}: Towards realistic high-resolution image blending.
\newblock In {\em Proceedings of the 27th ACM international conference on
  multimedia}, pages 2487--2495, 2019.

\bibitem{maskGuided_Manifold}
Mengyu Yang, David Rokeby, and Xavier Snelgrove.
\newblock {Mask-Guided Discovery of Semantic Manifolds in Generative Models}.
\newblock {\em 4th Workshop on Machine Learning for Creativity and Design at
  NeurIPS 2020}, 2020.

\bibitem{LSUN}
Fisher Yu, Ari Seff, Yinda Zhang, Shuran Song, Thomas Funkhouser, and Jianxiong
  Xiao.
\newblock {LSUN: Construction of a Large-scale Image Dataset using Deep
  Learning with Humans in the Loop}.
\newblock {\em arXiv preprint arXiv: 1506.03365}, 2015.

\bibitem{zhang2020deep}
Lingzhi Zhang, Tarmily Wen, and Jianbo Shi.
\newblock Deep image blending.
\newblock In {\em Proceedings of the IEEE/CVF Winter Conference on Applications
  of Computer Vision}, pages 231--240, 2020.

\bibitem{unreasonable_perceptual_metric}
Richard Zhang, Phillip Isola, Alexei~A Efros, Eli Shechtman, and Oliver Wang.
\newblock {The Unreasonable Effectiveness of Deep Features as a Perceptual
  Metric}.
\newblock In {\em Proceedings of the IEEE Conference on Computer Vision and
  Pattern Recognition}, pages 586--595, 2018.

\bibitem{InverseGraphics2020}
Yuxuan Zhang, Wenzheng Chen, Huan Ling, Jun Gao, Yinan Zhang, A. Torralba, and
  S. Fidler.
\newblock {Image GANs meet Differentiable Rendering for Inverse Graphics and
  Interpretable 3D Neural Rendering}.
\newblock {\em arXiv preprint arXiv: abs/2010.09125}, 2021.

\bibitem{SEAN2020}
Peihao Zhu, Rameen Abdal, Yipeng Qin, and Peter Wonka.
\newblock {SEAN: Image Synthesis with Semantic Region-Adaptive Normalization}.
\newblock {\em 2020 IEEE/CVF Conference on Computer Vision and Pattern
  Recognition}, pages 5103--5112, 2020.

\end{thebibliography}
}


\clearpage
\appendix

\section*{Appendix}

In addition to the appendix sections below, we provide our code publicly with the relevant instructions \footnote{ \scriptsize{
 \url{https://github.com/denabazazian/Dual-Domain-Synthesis}}}.
 
\section{Impact of pre-trained networks on perceptual loss}
We evaluate the impact of using the activations from different pre-trained  convolutional backbones on our model. We assess four backbone models namely VGG~\cite{VGG}, AlexNet~\cite{AlexNet}, SqueezeNet~\cite{Squeezenet}, ResNet~\cite{ResNet}, all pretrained on ImageNet.
The shape of the layers to calculate the losses in each backbone are included in Tables \ref{tab:VGG_layers} - \ref{tab:ResNet} respectively.

\begin{figure}[t]
\centering
\begin{center}
\setlength{\tabcolsep}{0.350pt}
\begin{tabular}{ c c c c c}
 
\small{\textit{Images}} &  
\includegraphics[width=0.09\textwidth]{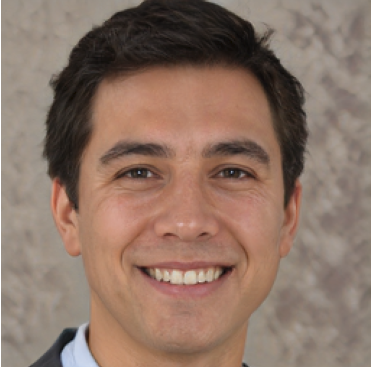}
 \makebox[0pt][r]{ \raisebox{0.0em}{\includegraphics[width=0.025\textwidth]{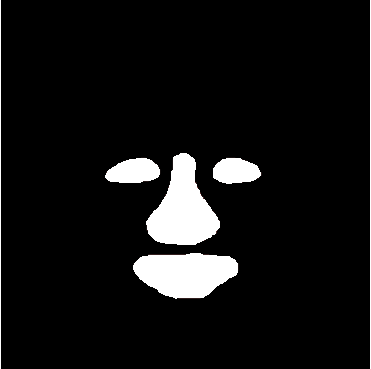} }}&
 \includegraphics[width=0.09\textwidth]{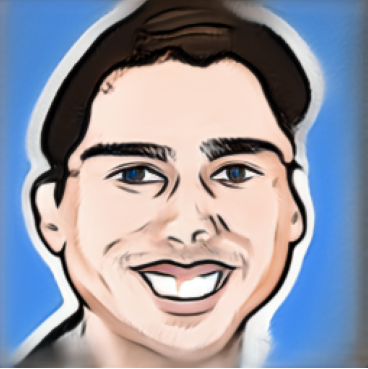}
 \makebox[0pt][r]{ \raisebox{0.0em}{\includegraphics[width=0.025\textwidth]{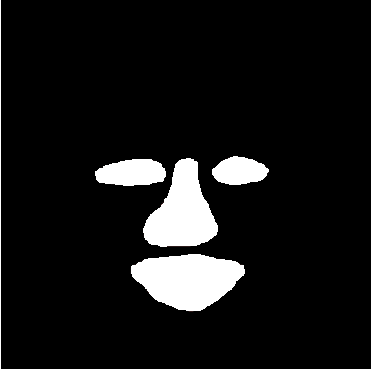} }}&
 \includegraphics[width=0.09\textwidth]{imgs/backbone/org_natural.png}
 \makebox[0pt][r]{ \raisebox{0.0em}{\includegraphics[width=0.025\textwidth]{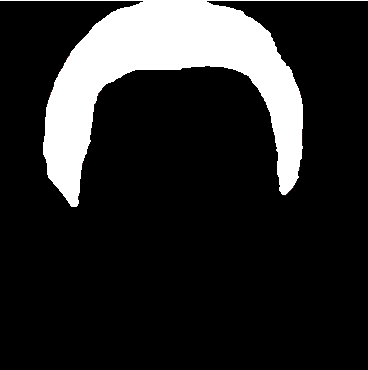} }}&
 \includegraphics[width=0.09\textwidth]{imgs/backbone/org_caricatur.png}
 \makebox[0pt][r]{ \raisebox{0.0em}{\includegraphics[width=0.025\textwidth]{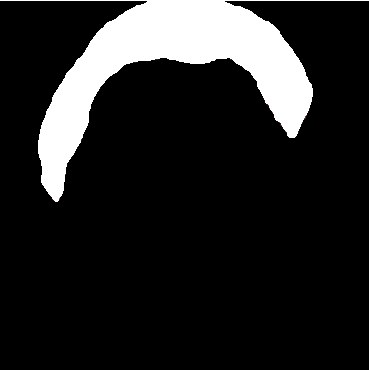} }}
 \\
\small{\textit{ResNet}} & \includegraphics[width=0.09\textwidth]{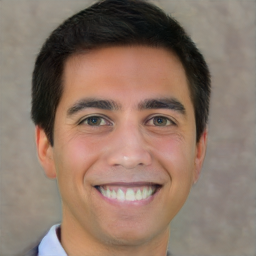} & \includegraphics[width=0.09\textwidth]{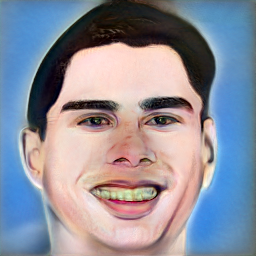} & \includegraphics[width=0.09\textwidth]{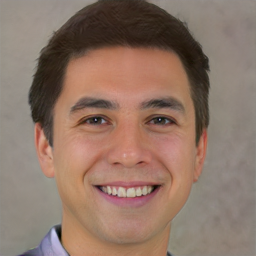} & \includegraphics[width=0.09\textwidth]{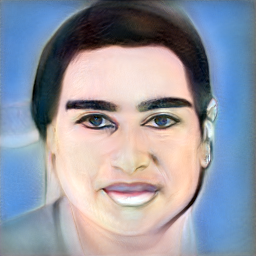}  
\\
  \small{\textit{SqueezeNet}} & \includegraphics[width=0.09\textwidth]{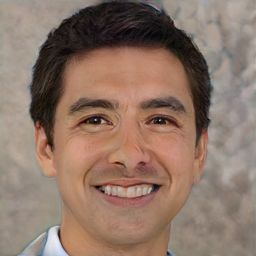} & \includegraphics[width=0.09\textwidth]{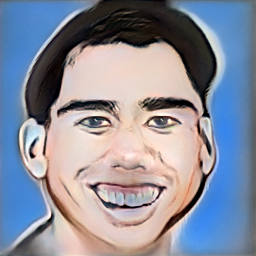} & \includegraphics[width=0.09\textwidth]{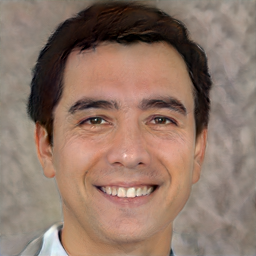} & \includegraphics[width=0.09\textwidth]{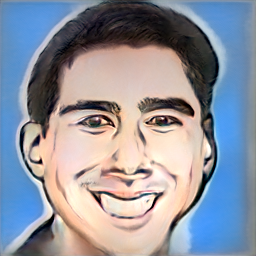} 
\\
 \small{\textit{AlexNet}} & \includegraphics[width=0.09\textwidth]{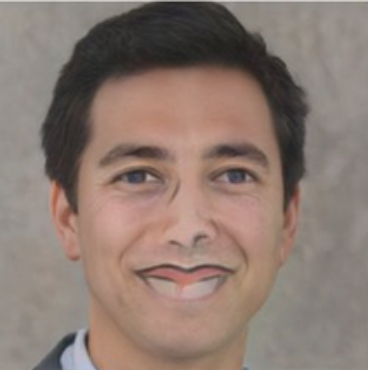} & \includegraphics[width=0.09\textwidth]{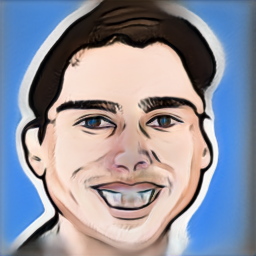} & \includegraphics[width=0.09\textwidth]{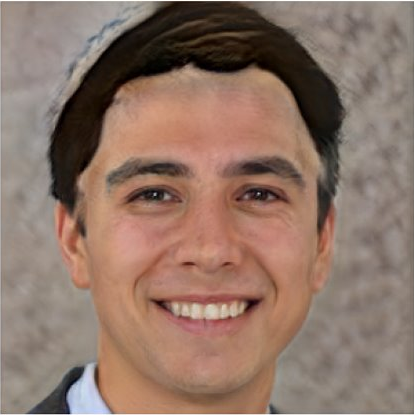} & \includegraphics[width=0.09\textwidth]{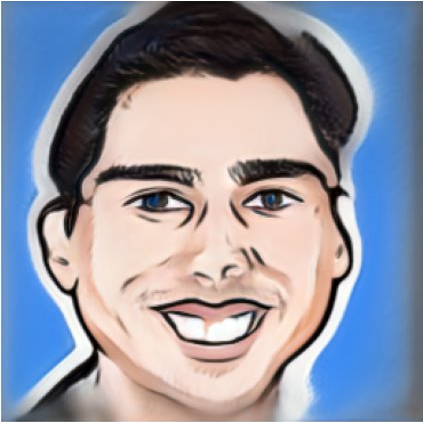} 
\\  
 \small{\textit{VGG}} & \includegraphics[width=0.09\textwidth]{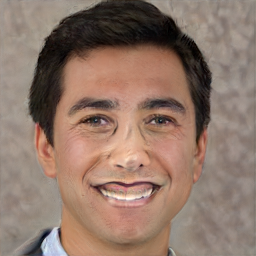} & \includegraphics[width=0.09\textwidth]{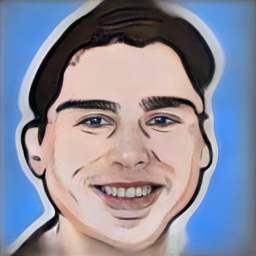} & \includegraphics[width=0.09\textwidth]{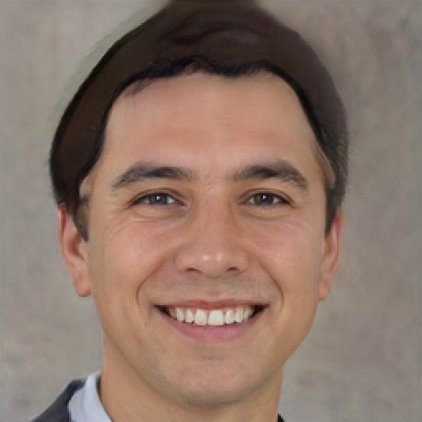} & \includegraphics[width=0.09\textwidth]{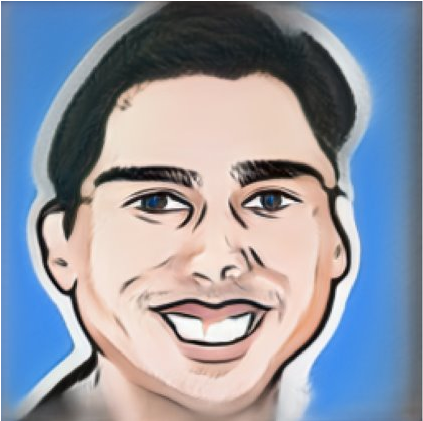} 
\\ 
\end{tabular}
\end{center}
\caption{The impact of four pre-trained backbones for computing perceptual loss. The first two columns uses the eye/nose/mouth segmentation model, while the last two columns uses the hair segmentation model.}
\label{fig:res_diff_network}
\end{figure}

Fig.~\ref{fig:res_diff_network} shows the qualitative results of the human face and caricature domains.
Table~\ref{tab:abl_res} shows the numerical results of this experiment on the hair segmentation model; all the metrics of this table are computed over the target domain. 
These experiments demonstrate that the pre-trained VGG~\cite{VGG} model  offers the best performance for the perceptual loss in faces. 
Previous work~\cite{image2stylegan, image2stylegan++} only applied pretrained VGG~\cite{VGG} model for perceptual loss, and in~\cite{unreasonable_perceptual_metric} the assumption for the perceptual loss was that the results from VGG~\cite{VGG} and AlexNet~\cite{AlexNet} are similar. We show qualitatively and quantitatively that some network backbones perform better than others. 

We also demonstrate that this is domain specific in Fig.~\ref{fig:horse} and Fig.~\ref{fig:car}. 
For both the horse and the car domains, AelxNet offers significantly better results than VGG.
This is an interesting observation given that the backbones are trained on the same data.
We stipulate that the architecture design has been motivated by different tasks.
Further exploration of the role of backbone in perceptual loss calculation is needed.




\begin{table}[h]
\caption{Comparison of the performance of four backbones for perceptual loss. The result are on the natural faces and caricature domains, where the segmentation model is trained to segment hair.}
    \centering
    \setlength{\tabcolsep}{2.1pt}
    \begin{tabular}{c  c |  c c c | c c c}
    \hline 
&  & \multicolumn{3}{c|}{ \scriptsize{$DDS_{car. \rightarrow nat.}$}} & \multicolumn{3}{c}{ \scriptsize{$DDS_{nat. \rightarrow car.}$}} \\
&   &  \small{FID$\downarrow$} & \small{SSIM$\uparrow$} & \small{PSNR$\uparrow$} &  \small{FID$\downarrow$} & \small{SSIM$\uparrow$} & \small{PSNR$\uparrow$} \\
\hline
\small{\textit{ResNet}}  & \multicolumn{1}{c|}{\small{$\mathcal{D}_t$}} & \small{95.66} & \small{0.55} & \small{28.89} & \small{98.71} & \small{0.47} & \small{27.93}
\\
\small{\textit{SqueezNet}}  & \multicolumn{1}{c|}{\small{$\mathcal{D}_t$}} & \small{89.48} & \small{0.62} & \small{29.15} & \small{94.29} & \small{0.52} & \small{28.05}
\\
\small{\textit{AlexNet}}  & \multicolumn{1}{c|}{\small{$\mathcal{D}_t$}} & \small{88.63} & \small{0.66} & \small{29.36} & \small{93.27} & \small{0.71} & \small{28.11}
\\
\small{\textit{VGG}}  & \multicolumn{1}{c|}{\small{$\mathcal{D}_t$}} & \small{ \textbf{87.28 }} & \small{\textbf{0.71}} & \small{ \textbf{29.58}} & \small{\textbf{76.96}} & \small{\textbf{0.72}} & \small{\textbf{28.39}} 
\\ \hline
  \end{tabular}
    \label{tab:abl_res}
\end{table}
\begin{figure}[th!]
\centering
\begin{center}
\setlength{\tabcolsep}{1.0pt}
\begin{tabular}{ c c | c c } 

\includegraphics[width=0.11\textwidth]{imgs/results/horse_car/horse/1_horse_org_source.png}
 \makebox[0pt][r]{ \raisebox{0.0em}{\includegraphics[width=0.030\textwidth]{imgs/results/horse_car/horse/1_horse_mask_source_0.png} }} &
\includegraphics[width=0.11\textwidth]{imgs/results/horse_car/horse/1_horse_org_target.png}
 \makebox[0pt][r]{ \raisebox{0.0em}{\includegraphics[width=0.030\textwidth]{imgs/results/horse_car/horse/1_horse_mask_target_0.png} }} &
\includegraphics[width=0.11\textwidth]{imgs/results/horse_car/horse/11_horse_org_source.png}
 \makebox[0pt][r]{ \raisebox{0.0em}{\includegraphics[width=0.030\textwidth]{imgs/results/horse_car/horse/11_horse_mask_source_0.png} }} &
\includegraphics[width=0.11\textwidth]{imgs/results/horse_car/horse/11_horse_org_target.png}
 \makebox[0pt][r]{ \raisebox{0.0em}{\includegraphics[width=0.030\textwidth]{imgs/results/horse_car/horse/11_horse_mask_target_0.png} }}
\\
\includegraphics[width=0.11\textwidth]{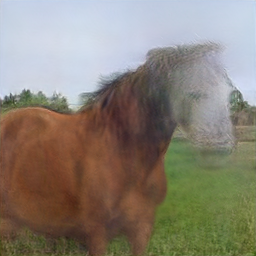}&
\includegraphics[width=0.11\textwidth]{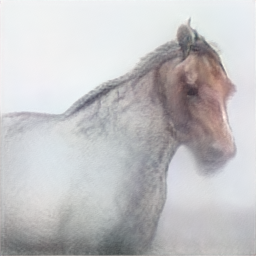}&
\includegraphics[width=0.11\textwidth]{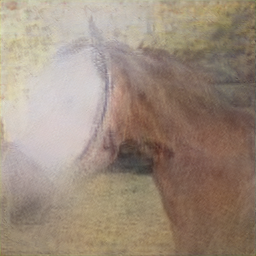}&
\includegraphics[width=0.11\textwidth]{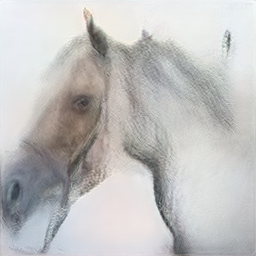}
\\
\includegraphics[width=0.11\textwidth]{imgs/results/horse_car/horse/1_horse_AlexNet_003002_source.png}&
\includegraphics[width=0.11\textwidth]{imgs/results/horse_car/horse/1_horse_AlexNet_003002_target.png}&
\includegraphics[width=0.11\textwidth]{imgs/results/horse_car/horse/11_horse_AlexNet_003002_source.png}&
\includegraphics[width=0.11\textwidth]{imgs/results/horse_car/horse/11_horse_AlexNet_003002_target.png}
\\ 

\end{tabular}
\end{center}
\vspace*{-12pt}
\caption{DDS results on horse domains. Odd columns are from natural horse domain, and even columns are from sketch horse domain. Segmentation masks are Incorporated.  
Second and third rows show the corresponding dual-domain images using the perceptual loss from the $conv$ layers of VGG~\cite{VGG} and AlexNet~\cite{AlexNet} respectively. }
\label{fig:horse}
\end{figure}


\begin{figure}[th!]
\centering
\begin{center}
\setlength{\tabcolsep}{0.6pt}
\begin{tabular}{ c c | c c }

 \includegraphics[width=0.11\textwidth]{imgs/results/horse_car/car/19_car_abondoned_org_source.png}
 \makebox[0pt][r]{ \raisebox{0.5em}{\includegraphics[width=0.030\textwidth]{imgs/results/horse_car/car/19_car_abondoned_mask_source_0.png} }} &
\includegraphics[width=0.11\textwidth]{imgs/results/horse_car/car/19_car_abondoned_org_target.png}
 \makebox[0pt][r]{ \raisebox{0.0em}{\includegraphics[width=0.030\textwidth]{imgs/results/horse_car/car/19_car_abondoned_mask_target_0.png} }}&
\includegraphics[width=0.11\textwidth]{imgs/results/horse_car/car/16_car_abondoned_org_source.png}
 \makebox[0pt][r]{ \raisebox{0.5em}{\includegraphics[width=0.030\textwidth]{imgs/results/horse_car/car/16_car_abondoned_mask_source_0.png} }} &
\includegraphics[width=0.11\textwidth]{imgs/results/horse_car/car/16_car_abondoned_org_target.png}
 \makebox[0pt][r]{ \raisebox{0.0em}{\includegraphics[width=0.030\textwidth]{imgs/results/horse_car/car/16_car_abondoned_mask_target_0.png} }}

\\
\includegraphics[width=0.11\textwidth]{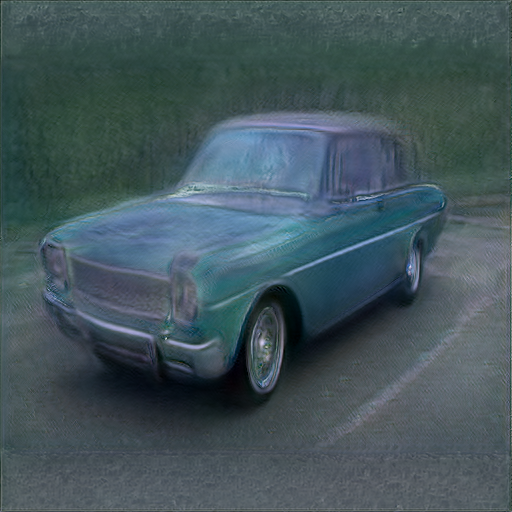}&
\includegraphics[width=0.11\textwidth]{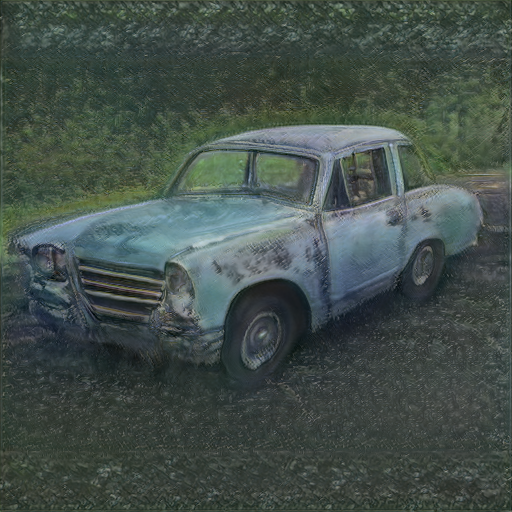}&
\includegraphics[width=0.11\textwidth]{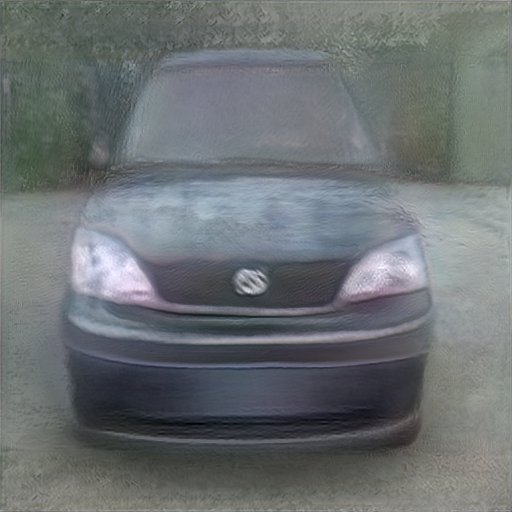}&
\includegraphics[width=0.11\textwidth]{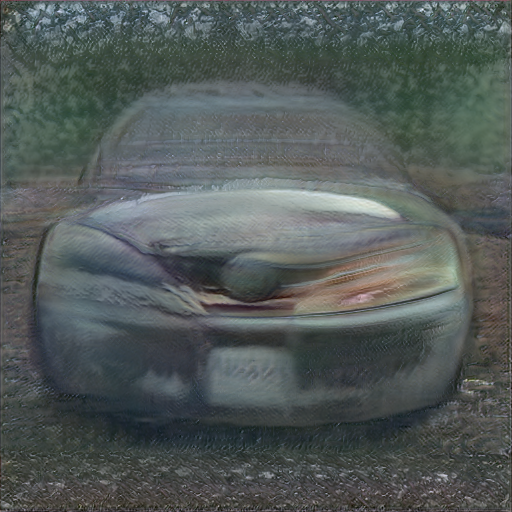}
\\
\includegraphics[width=0.11\textwidth]{imgs/results/horse_car/car/19_car_abondoned_003002_source.png}&
\includegraphics[width=0.11\textwidth]{imgs/results/horse_car/car/19_car_abondoned_003002_target.png}&
\includegraphics[width=0.11\textwidth]{imgs/results/horse_car/car/16_car_abondoned_003002_source.png}&
\includegraphics[width=0.11\textwidth]{imgs/results/horse_car/car/16_car_abondoned_003002_target.png}
\\
 
\end{tabular}
\end{center}
\vspace*{-12pt}
\caption{DDS results on car domains. Odd columns are standard car images, and even columns are abandoned cars. Segmentation masks are incorporated. 
Second and third row show the DDS results using the perceptual loss from the $conv$ layers of VGG~\cite{VGG} and AlexNet~\cite{AlexNet} respectively.}
\label{fig:car}
\end{figure}

 \section{Dual-Domain Synthesis vs. Image Blending}
\begin{figure}[h!]   
 \centering
 \begin{center}
 \setlength{\tabcolsep}{0.001pt}
 \begin{tabular}{ c c c c c} 


 \includegraphics[width=0.090\textwidth]{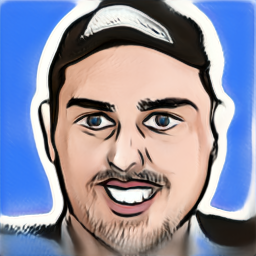}
 \makebox[0pt][r]{ \raisebox{0em}{\includegraphics[width=0.020\textwidth]{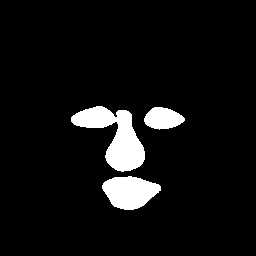} }} &
 \includegraphics[width=0.090\textwidth]{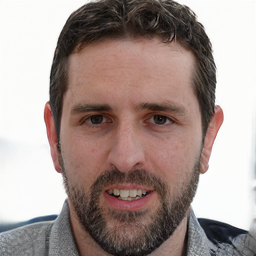}
 \makebox[0pt][r]{ \raisebox{0em}{\includegraphics[width=0.020\textwidth]{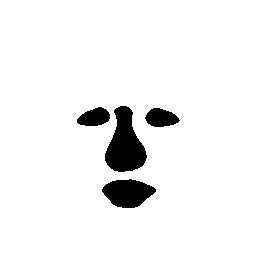} }} &
 \includegraphics[width=0.090\textwidth]{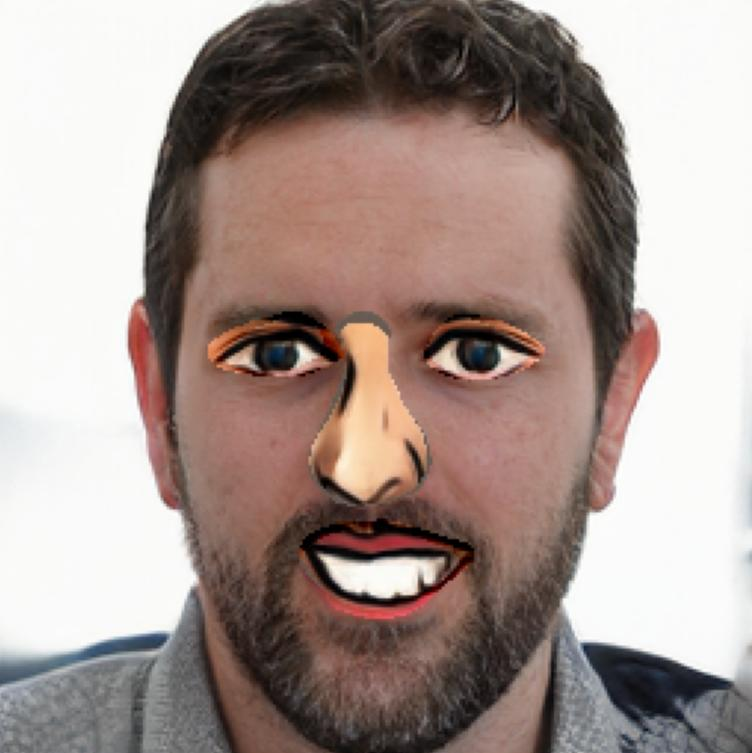} &
 \includegraphics[width=0.090\textwidth]{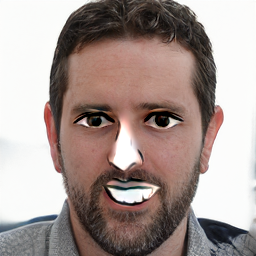} &
 \includegraphics[width=0.090\textwidth]{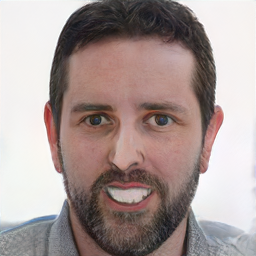} 
 \\
 \includegraphics[width=0.090\textwidth]{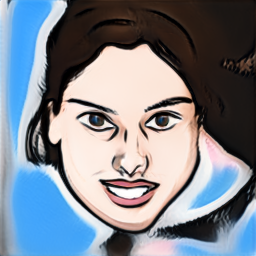}
 \makebox[0pt][r]{ \raisebox{0em}{\includegraphics[width=0.020\textwidth]{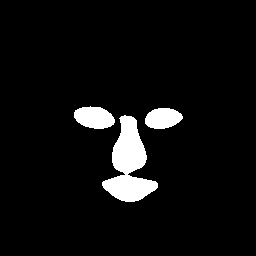} }} &
 \includegraphics[width=0.090\textwidth]{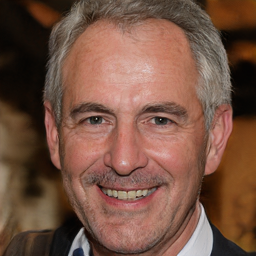}
 \makebox[0pt][r]{ \raisebox{0em}{\includegraphics[width=0.020\textwidth]{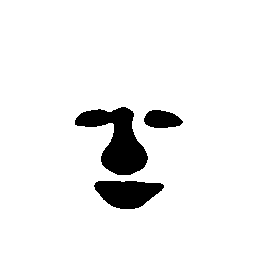} }} &
 \includegraphics[width=0.090\textwidth]{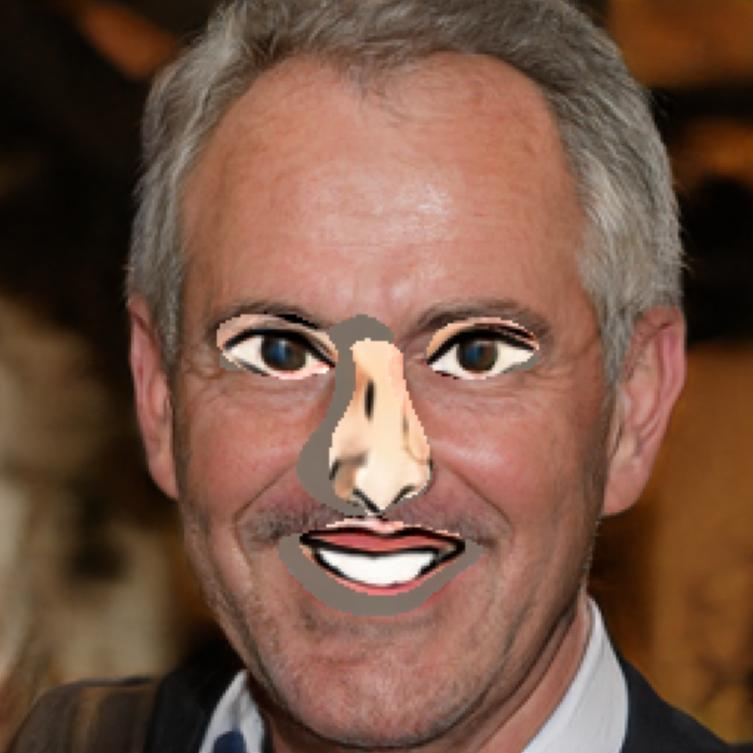} &
 \includegraphics[width=0.090\textwidth]{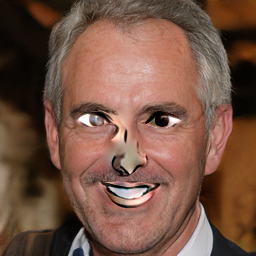} &
 \includegraphics[width=0.090\textwidth]{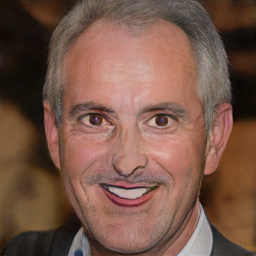}

 \end{tabular}
 \end{center}
 \vspace*{-15pt}
 \caption{Comparing~\cite{sofiiuk2021foreground} (col3) and \cite{zhang2020deep} (col4) to Dual-Domain (col5), for paired (top) and unpaired (bottom) examples.}
 \label{fig:comparingWCT1}
 \end{figure}

 To demonstrate the difference of our DDS with image blending approaches, we provide a comparison in Fig.~\ref{fig:comparingWCT1} between Dual-Domain and recent Image Blending works (using the public codes of~\cite{zhang2020deep,sofiiuk2021foreground}).
 Blending does not achieve better or even comparable results.

\section{DDS from few-shotGAN domains}
\begin{figure*}
\centering
\begin{center}
\setlength{\tabcolsep}{0.010pt}
\begin{tabular}{ c c | c c | c c }

 \includegraphics[width=0.15\textwidth]{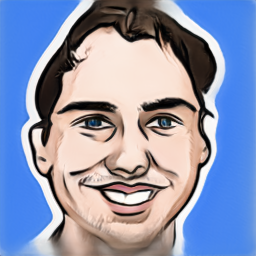}
 \makebox[0pt][r]{ \raisebox{0.0em}{\includegraphics[width=0.030\textwidth]{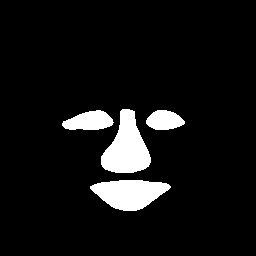} }} &

 \includegraphics[width=0.15\textwidth]{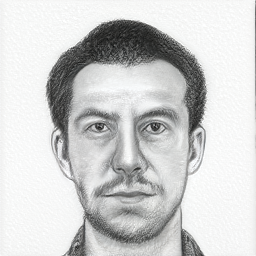}
 \makebox[0pt][r]{ \raisebox{0.0em}{\includegraphics[width=0.030\textwidth]{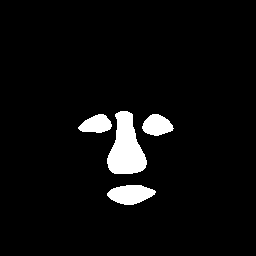} }} &

 \includegraphics[width=0.15\textwidth]{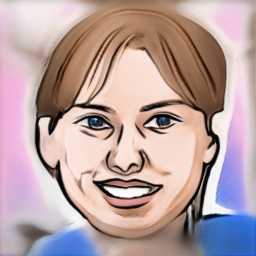}
 \makebox[0pt][r]{ \raisebox{0.0em}{\includegraphics[width=0.030\textwidth]{imgs/results/supplementary/caricature_sketch/1_mask_source_0.png} }} &

 \includegraphics[width=0.15\textwidth]{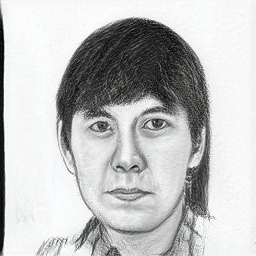}
 \makebox[0pt][r]{ \raisebox{0.0em}{\includegraphics[width=0.030\textwidth]{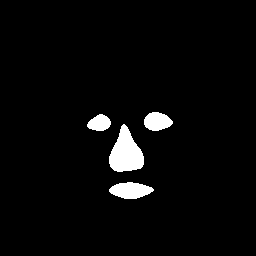} }} &
 

 
  \includegraphics[width=0.15\textwidth]{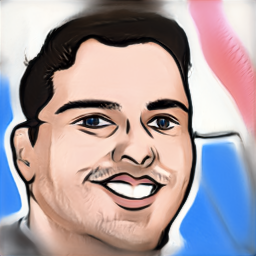}
 \makebox[0pt][r]{ \raisebox{0.0em}{\includegraphics[width=0.030\textwidth]{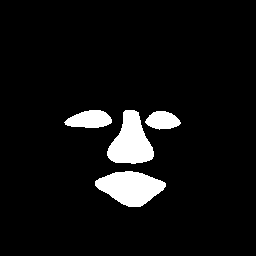} }} &

 \includegraphics[width=0.15\textwidth]{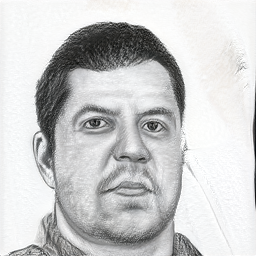}
 \makebox[0pt][r]{ \raisebox{0.0em}{\includegraphics[width=0.030\textwidth]{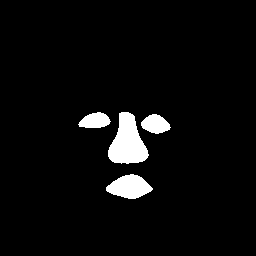} }} 
 \\
\includegraphics[width=0.15\textwidth]{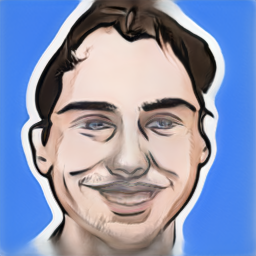} &
 \includegraphics[width=0.15\textwidth]{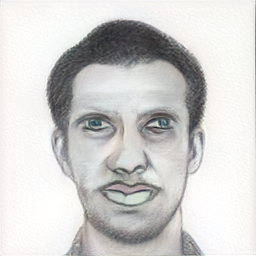}&
 \includegraphics[width=0.15\textwidth]{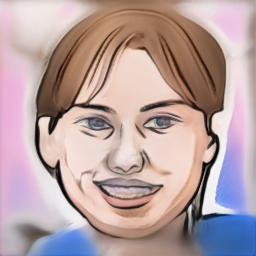} &
 \includegraphics[width=0.15\textwidth]{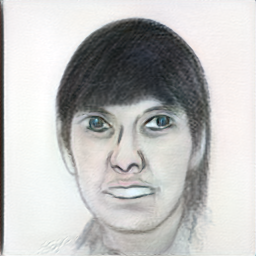} &
\includegraphics[width=0.15\textwidth]{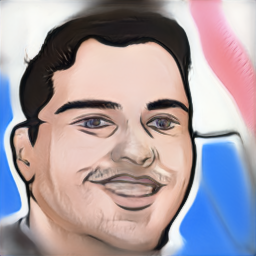} &
 \includegraphics[width=0.15\textwidth]{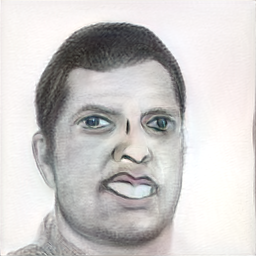}

\end{tabular}
\end{center}
\caption{DDS results on caricature and sketch domains. The first row show a pair of images from the domains of caricature and sketch. The eyes/nose/mouth segmentation model used in these experiments, and all the correspondences masks are shown in the first row. The second row shows the dual-domain images containing the features of caricature and sketch based on the segmentation model. The changes in the shape of the nose and mouth and colour of the eyes can be observed from the dual-domain images.  }
\label{fig:car_sket}
\end{figure*}

We performed further experiments on synthesising images which contain features from both the caricature and sketch domains. 
Note that both GANs are trained using few-shots, adapted independently from the natural face StyleGAN.
Fig.~\ref{fig:car_sket} shows dual-domain images based on the integrating caricature and sketch features.

\section{Final vs. intermediate latent space editing} 
\label{sec: latent_analysis}
\begin{figure*}
\centering
\begin{center}
\setlength{\tabcolsep}{0.6pt}
\begin{tabular}{ c c | c c | c c}

\includegraphics[width=0.15\textwidth]{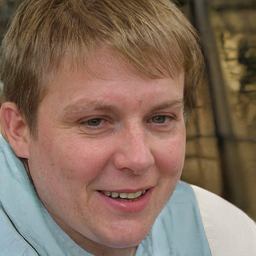} 
\makebox[0pt][r]{ \raisebox{0em}{\includegraphics[width=0.030\textwidth]{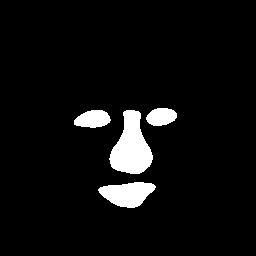} }}&
\includegraphics[width=0.15\textwidth]{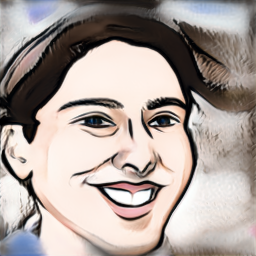} 
\makebox[0pt][r]{ \raisebox{0em}{\includegraphics[width=0.030\textwidth]{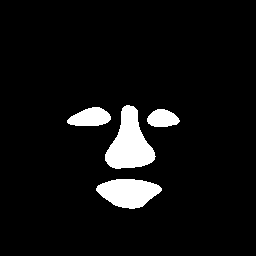} }} &

 
  \includegraphics[width=0.15\textwidth]{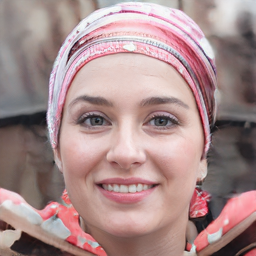}
 \makebox[0pt][r]{ \raisebox{0.0em}{\includegraphics[width=0.030\textwidth]{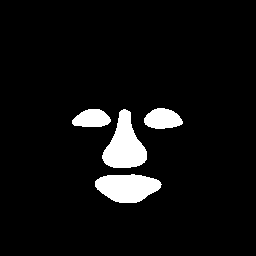} }} &
 
  \includegraphics[width=0.15\textwidth]{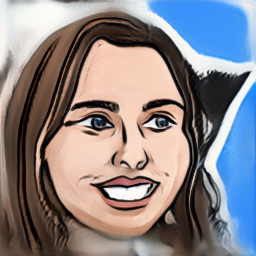}
 \makebox[0pt][r]{ \raisebox{0.0em}{\includegraphics[width=0.030\textwidth]{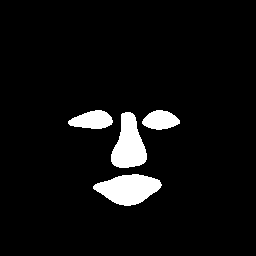} }} &
 
   \includegraphics[width=0.15\textwidth]{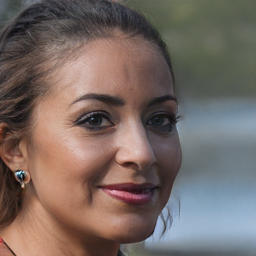}
 \makebox[0pt][r]{ \raisebox{0.0em}{\includegraphics[width=0.030\textwidth]{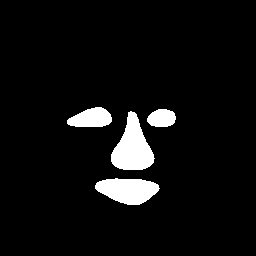} }} &
 
  \includegraphics[width=0.15\textwidth]{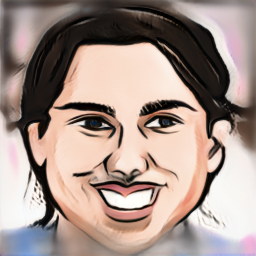}
 \makebox[0pt][r]{ \raisebox{0.0em}{\includegraphics[width=0.030\textwidth]{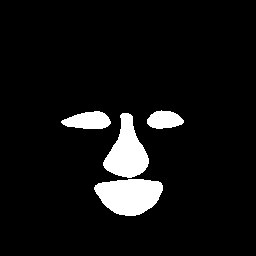} }} 
\\
 \includegraphics[width=0.15\textwidth]{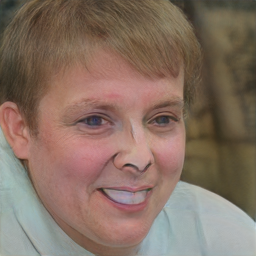} &
\includegraphics[width=0.15\textwidth]{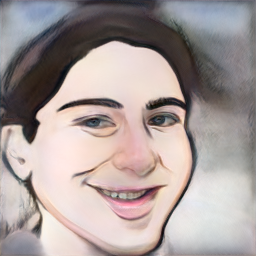} &
  \includegraphics[width=0.15\textwidth]{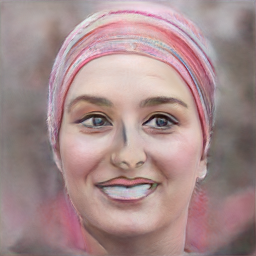} &
 \includegraphics[width=0.15\textwidth]{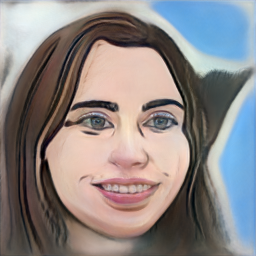} &
  \includegraphics[width=0.15\textwidth]{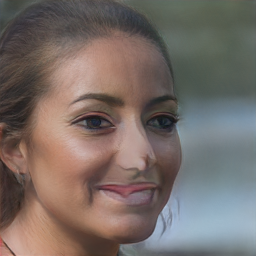} &
 \includegraphics[width=0.15\textwidth]{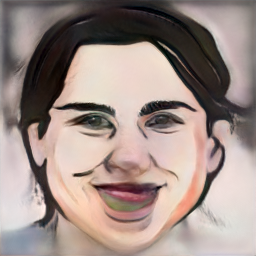} 
\\
  \includegraphics[width=0.15\textwidth]{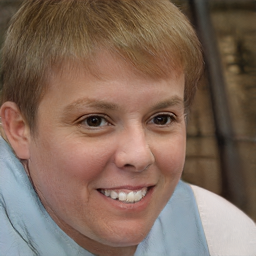} &
 \includegraphics[width=0.15\textwidth]{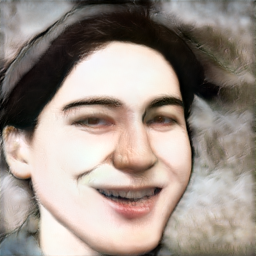} &
  \includegraphics[width=0.15\textwidth]{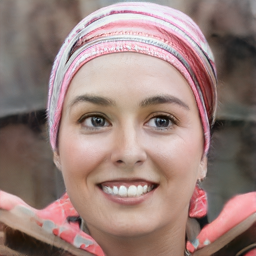} &
 \includegraphics[width=0.15\textwidth]{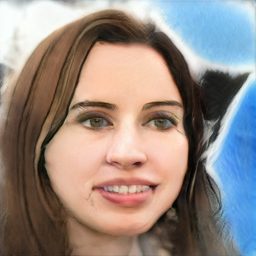} &
  \includegraphics[width=0.15\textwidth]{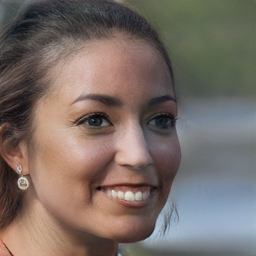} &
 \includegraphics[width=0.15\textwidth]{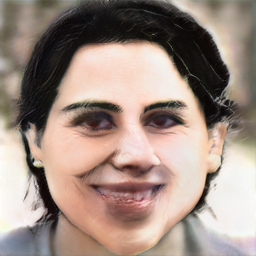}

\end{tabular}
\end{center}
\caption{Comparison of the results of using the intermediate latent activations. First row shows unpaired images of two domains, masks are incorporated. Second row shows the DDS results, where each image maintain distinct features from the two domains. Third row shows the results when we use the intermediate latent space features for computing the perceptual loss. While image are integrated, they do not maintain the distinctness of the two domains. For example, in the first image (row 3, col 1) the output is purely a natural face.}
\label{fig:latent}
\end{figure*}

In our results, we used the synthesised image from the last layer of StyleGAN.
Instead, intermediate representations can be used to accommodate unpaired images of varying poses.
We experiment with using the concatenation of hidden intermediate layer activations. In this experiment for the images with resolution $256 \times 256$, we get the $13$ layers of features as described in Table~\ref{tab:styleganW}. Hence, for computing the perceptual loss instead of passing the images to $conv$ layers of \emph{VGG}, we directly use the activations of intermediate latent space of styleGAN from both the target and source domains. Fig.~\ref{fig:latent} demonstrates a comparison of the results when we apply DDS framework and when we use the intermediate latent activations.

While intermediate activations can accommodate variations in pose, they cannot maintain the features from two domains. 

\begin{table}[h!]
    \centering
     \caption{Shape of the intermediate StyleGAN features. }
    \begin{tabular}{c|c}
       $\mathbf{w} \in \mathcal{W}$  &  $shape$ \\ \hline
        $\mathbf{w}_0$ & $[512 \times 4 \times 4]$ \\
        $\mathbf{w}_1$ & $[512 \times 8 \times 8]$ \\
        $\mathbf{w}_2$ & $[512 \times 8 \times 8]$ \\
        $\mathbf{w}_3$ & $[512 \times 16 \times 16]$ \\
        $\mathbf{w}_4$ & $[512 \times 16 \times 16]$ \\
        $\mathbf{w}_5$ & $[512 \times 32 \times 32]$ \\
        $\mathbf{w}_6$ & $[512 \times 32 \times 32]$ \\
        $\mathbf{w}_7$ & $[512 \times 64 \times 64]$ \\
        $\mathbf{w}_8$ & $[512 \times 64 \times 64]$ \\
        $\mathbf{w}_9$ & $[256 \times 128 \times 128]$ \\
        $\mathbf{w}_{10}$ & $[256 \times 128 \times 128]$ \\
        $\mathbf{w}_{11}$ & $[128 \times 256 \times 256]$ \\
        $\mathbf{w}_{12}$ & $[128 \times 256 \times 256]$ \\
    \end{tabular}
   
    \label{tab:styleganW}

    \centering
        \caption{Shape of the \emph{VGG-{16}} layers that used for perceptual loss. }
    \begin{tabular}{c|c}
         $Layer$ & $shape$ \\ \hline
        $conv_{1\_1}$ & $[64 \times 256 \times 256]$ \\
        $conv_{1\_2}$ & $[64 \times 256 \times 256]$ \\
        $conv_{2\_2}$ & $[256 \times 64 \times 64]$ \\
        $conv_{3\_3}$ & $[512 \times 32 \times 32]$
    \end{tabular}

    \label{tab:VGG_layers}

    \centering
        \caption{Shape of the \emph{AlexNet} layers that used for perceptual loss. }
    \begin{tabular}{c|c}
         $Layer$ & $shape$ \\ \hline
        $relu_{1}$ & $[64 \times 55 \times 55]$ \\
        $relu_{2}$ & $[192 \times 27 \times 27]$ \\
        $relu_{3}$ & $[384 \times 13 \times 13]$ \\
        $relu_{4}$ & $[256 \times 13 \times 13]$ \\
        $relu_{5}$ & $[256 \times 13 \times 13]$ 
    \end{tabular}

    \label{tab:AlexNet}

    \centering
        \caption{Shape of the \emph{SqueezeNet} layers that used for perceptual loss. }
    \begin{tabular}{c|c}
         $Layer$ & $shape$ \\ \hline
        $relu_{1}$ & $[64 \times 127 \times 127]$ \\
        $relu_{2}$ & $[128 \times 63 \times 63]$ \\
        $relu_{3}$ & $[256 \times 31 \times 31]$ \\
        $relu_{4}$ & $[384 \times 15 \times 15]$ \\
        $relu_{5}$ & $[384 \times 15 \times 15]$ \\
        $relu_{6}$ & $[512 \times 15 \times 15]$ \\
        $relu_{7}$ & $[512 \times 15 \times 15]$ 
    \end{tabular}

    \label{tab:SqueezeNet}

    \centering
        \caption{Shape of the \emph{ResNet-{18}} layers that used for perceptual loss. }
    \begin{tabular}{c|c}
         $Layer$ & $shape$ \\ \hline
        $relu_{1}$ & $[64 \times 128 \times 128]$ \\
        $conv_{2}$ & $[64 \times 64 \times 64]$ \\
        $conv_{3}$ & $[128 \times 32 \times 32]$ \\
        $conv_{4}$ & $[256 \times 16 \times 16]$ \\
        $conv_{5}$ & $[512 \times 8 \times 8]$
    \end{tabular}

    \label{tab:ResNet}
\end{table}

\section {Sketch horse data} 
\label{sec: sketch_horse_data}
\begin{figure*}[h!] 
\centering
\begin{center}
\setlength{\tabcolsep}{0.9pt}
\begin{tabular}{ c c c c c }

\includegraphics[width=0.19\textwidth]{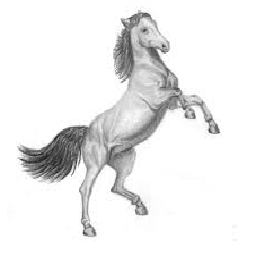} &
\includegraphics[width=0.19\textwidth]{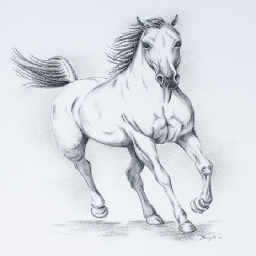} &
\includegraphics[width=0.19\textwidth]{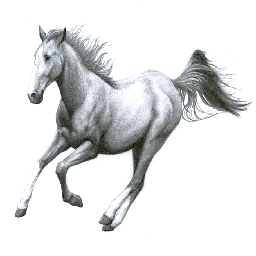} &
\includegraphics[width=0.19\textwidth]{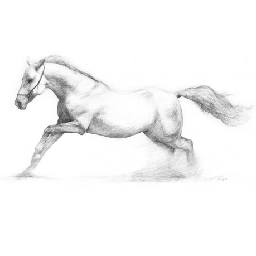} &
\includegraphics[width=0.19\textwidth]{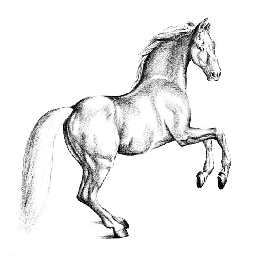} 
\\
\includegraphics[width=0.19\textwidth]{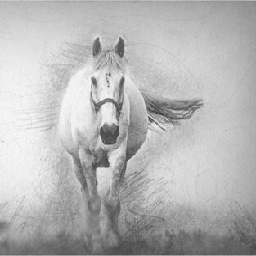} &
\includegraphics[width=0.19\textwidth]{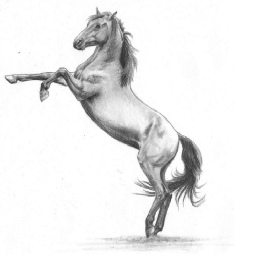} &
\includegraphics[width=0.19\textwidth]{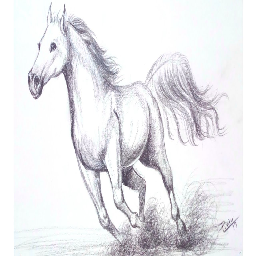} &
\includegraphics[width=0.19\textwidth]{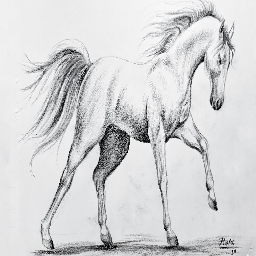} &
\includegraphics[width=0.19\textwidth]{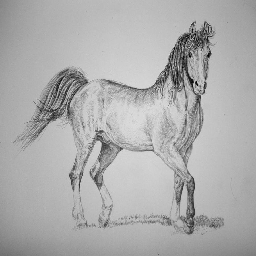} 

\end{tabular}

\caption{Training samples for generating sketch horse images. }
\label{fig:sketch_horse_training_data}
\end{center}
\end{figure*}

Fig.~\ref{fig:sketch_horse_training_data} 
shows the images of sketch horses that used for training few-shotGAN. 

\end{document}